\documentclass[10pt,doublecolumn]{IEEEtran}
\usepackage{amsmath,amsthm,amsfonts,amssymb,bm,mathrsfs}
\usepackage{graphicx,subfigure}
\usepackage{cite}
\usepackage{algorithm}
\usepackage{algorithmic}
\usepackage{multirow,booktabs,threeparttable}
\usepackage{tablefootnote}
\usepackage{color}
\usepackage{verbatim}
\usepackage{balance}
\graphicspath{{fig/}}

\theoremstyle{remark}

\theoremstyle{plain}

\begin{document}
\title{Stigmergic Independent Reinforcement Learning for Multi-Agent Collaboration}

\author{Xing Xu\textsuperscript{\rm 1}, Rongpeng Li\textsuperscript{\rm 1*}, Zhifeng Zhao\textsuperscript{\rm 2}, Honggang Zhang\textsuperscript{\rm 1}\\ 
\textsuperscript{\rm 1}Zhejiang University, \textsuperscript{\rm 2}Zhejiang Lab\\
\{hsuxing, lirongpeng, honggangzhang\}@zju.edu.cn, zhaozf@zhejianglab.com
}

\maketitle
\thispagestyle{empty}
\begin{abstract}
With the rapid evolution of wireless mobile devices, there emerges an increased need to design effective collaboration mechanisms between intelligent agents, so as to gradually approach the final collective objective through continuously learning from the environment based on their individual observations. In this regard, independent reinforcement learning (IRL) is often deployed in multi-agent collaboration to alleviate the problem of a non-stationary learning environment. However, behavioral strategies of intelligent agents in IRL can only be formulated upon their local individual observations of the global environment, and appropriate communication mechanisms must be introduced to reduce their behavioral localities. In this paper, we address the problem of communication between intelligent agents in IRL by jointly adopting mechanisms with two different scales. For the large scale, we introduce the stigmergy mechanism as an indirect communication bridge between independent learning agents, and carefully design a mathematical method to indicate the impact of digital pheromone. For the small scale, we propose a conflict-avoidance mechanism between adjacent agents by implementing an additionally embedded neural network to provide more opportunities for participants with higher action priorities. In addition, we present a federal training method to effectively optimize the neural network of each agent in a decentralized manner. Finally, we establish a simulation scenario in which a number of mobile agents in a certain area move automatically to form a specified target shape. Extensive simulations demonstrate the effectiveness of our proposed method.
\end{abstract}


\section{Introduction}
With the rapid development of mobile wireless communication and Internet of Things technologies, many scenarios have arisen in which collaboration between intelligent agents is required, such as in the deployment of unmanned aerial vehicles (UAVs) \cite{Schwarzrock2018Solving,Parunak2004Digital,Cimino2016Combining}, distributed control in the field of industry automation \cite{Hadeli2003Self,Hadeli2004Multi,Werfel2006Extended}, and mobile crowd sensing and computing (MCSC) \cite{Guo2016Mobile,Alfeo2017Stigmergy}. In these scenarios, traditional centralized control methods are usually impractical due to limited computational resources and the demand for ultra-low latency and ultra-high reliability. As an alternative, multi-agent collaboration technologies can be used in these scenarios to relieve the pressure on the centralized controller.

Guiding autonomous agents to act optimally through trial-and-error interaction with the corresponding environment is a primary goal in the field of artificial intelligence, and is regarded as one of the most important objectives of reinforcement learning (RL) \cite{Tesauro1995Temporal,Kohl2004Policy,Arulkumaran2017Deep}. Recently, deep RL (DRL), which combines RL and deep neural networks, has improved the ability to obtain information from high-dimensional input, such as high-resolution images, and has demonstrated extraordinary learning ability across a wide array of tasks\cite{Volodymyr2015Human}. There are a number of advanced DRL algorithms that can direct a single agent to improve its behavioral policy through continuously learning from the environment\cite{Mnih2016Asynchronous,Silver2014Deterministic}. However, the extension of single-agent DRL to multi-agent DRL is not straightforward, and many challenging problems remain to be solved in the application of multi-agent RL (MARL) \cite{Busoniu2008A,Littman1994Markov}. In particular, in a completely distributed multi-agent system (MAS), each agent is usually limited to partially observe the global environment, and its learning process following this local observation can thus be non-stationary, as other agents' behavioral policies may change temporally. In addition, the assignment of an individual reward is another challenging problem, as there is only one global reward for feedback in most cases. As an alternative, independent RL (IRL) has been proposed to alleviate the problem of a non-stationary learning environment, where each agent undergoes an independent learning process with only self-related sensations \cite{Matignon2012Independent}.

In IRL, most behavioral policies learned by intelligent agents are self-centered, aiming to maximize their own interests. Thus, the target of mutual communication is to integrate these individual behaviors effectively for the same task. Several studies have attempted to solve the problem of mutual communication in IRL. Foerster \emph{et al.} \cite{Foerster2016Learning} proposed differentiable inter-agent learning (DIAL), in which an additional communication action is added to the action set of each agent. In addition to the selection of the current action, a piece of inter-agent message is also generated and sent to other agents through a specified communication channel. Thereafter, a back-propagated error from the receiver of the communication messages is sent back to the sender to regulate the communication action. However, this type of communication channel exists between all pairs of independent learning agents, making DIAL very complex as the number of agents increases. Tan \cite{Tan1993Multi} tested to share different types of messages to coordinate intelligent agents on the basis of independent Q-learning (e.g., sensations, episodes, and learned policies). Despite its improvement in the final performance, the method has several over-optimistic assumptions that limit its potential application. For example, given the limited resources in a harsh environment, it is usually impractical for a single agent to transmit large messages in mobile wireless scenarios. As an improvement, Mao \emph{et al.} \cite{Mao2017ACCNet} proposed to utilize a coordinator network to aggregate compressed local messages and then share them among all agents. Because the shared messages contain joint information from all agents, the expected result of this design is to make each agent act optimally considering the other agents' behaviors. However, it is difficult to obtain a well-trained coordinator and local networks. In summary, appropriate communication mechanisms must be introduced between independent learning agents to reduce their behavioral localities \cite{Mataric1998Using}. 

As another approach, the concept of stigmergy was first introduced by French entomologist Pierre-Paul Grass\`e in the 1950s when studying the behavior of social insects \cite{Heylighen2016Stigmergy1,Heylighen2016Stigmergy2}. Recently, stigmergy has experienced rapid diffusion across a large number of application domains together with the popularization of distributed computing, collective intelligence, and broadband Internet \cite{Schoonderwoerd1996Ant,Caro1999AntNet,Dorigo2000Ant,Maru2017QoE}. In particular, stigmergy has demonstrated advantages in various scenarios requiring distributed control where the generation of messages is closely related to the environment, space, and time, such as the management of traffic lights \cite{Takahashi2012A,Kanamori2013Evaluation}. A key component of stigmergy is called the medium, which acts as the information aggregator in multi-agent collaboration. Benefiting from the existence of a medium, effective stigmergic interaction can be established between agents and their surrounding environment, and distributed agents can also interact with each other indirectly to reduce their behavioral localities. Therefore, stigmergy is a potential technique to solve the problem of mutual communication in IRL.
 
In addition to reap the large-scale merits of stigmergy, we also propose a conflict-avoidance mechanism executed at smaller scales between adjacent agents to further reduce their behavioral localities. With this mechanism, we evaluate and assign a corresponding action priority to each agent, and a larger number of decision-making opportunities are provided for agents with higher action priorities. In particular, the action priority value is efficiently calculated by an additionally embedded neural network within each agent. Furthermore, to synchronously optimize the neural network of each agent, we apply a federal training method along with average optimization by improving the asynchronous advantage actor-critic (A3C) algorithm \cite{Mnih2016Asynchronous}. We summarize all the aforementioned techniques and propose the stigmergic IRL (SIRL) algorithm. Based on the simulation scenario in \cite{Xing2019Brain}, in which a number of mobile agents move automatically to form a specified target shape, we examine the effectiveness of the proposed SIRL algorithm through an in-depth performance comparison with other available methods. 

Focusing on the SIRL algorithm, the contributions of this paper can be summarized as follows:

\begin{itemize}
	\item First, we introduce the stigmergy mechanism into MARL, and provide an effective cooperative algorithm. We also demonstrate that the stigmergy mechanism can decompose a global objective into small tasks that can be more efficiently perceived by individual agents.
	
	\item Second, we propose a conflict-avoidance mechanism to further reduce the behavioral localities of agents, whose foundation is an additionally embedded neural network in each agent. 
	
	\item Third, we provide a federal training method by enhancing the A3C algorithm to synchronously optimize the neural network of each independent learning agent in MAS.
\end{itemize}

The remainder of this paper is organized as follows. In Section \uppercase\expandafter{\romannumeral2}, we discuss the related work from the perspective of combining the stigmergy mechanism and MARL, and clarify the novelty of our work. In Section \uppercase\expandafter{\romannumeral3}, we present the system framework followed by a description of the stigmergy mechanism and the proposed conflict-avoidance mechanism. Finally, we introduce the federal training method. In Section \uppercase\expandafter{\romannumeral4}, we mathematically analyze the details of the proposed SIRL algorithm. In Section \uppercase\expandafter{\romannumeral5}, we describe the simulation scenario, compare the numerical simulation results, and present key insights into these results. Finally, in Section \uppercase\expandafter{\romannumeral6}, we conclude the paper. 

\begin{figure*}
	\centering
	\includegraphics[width=0.7\textwidth]{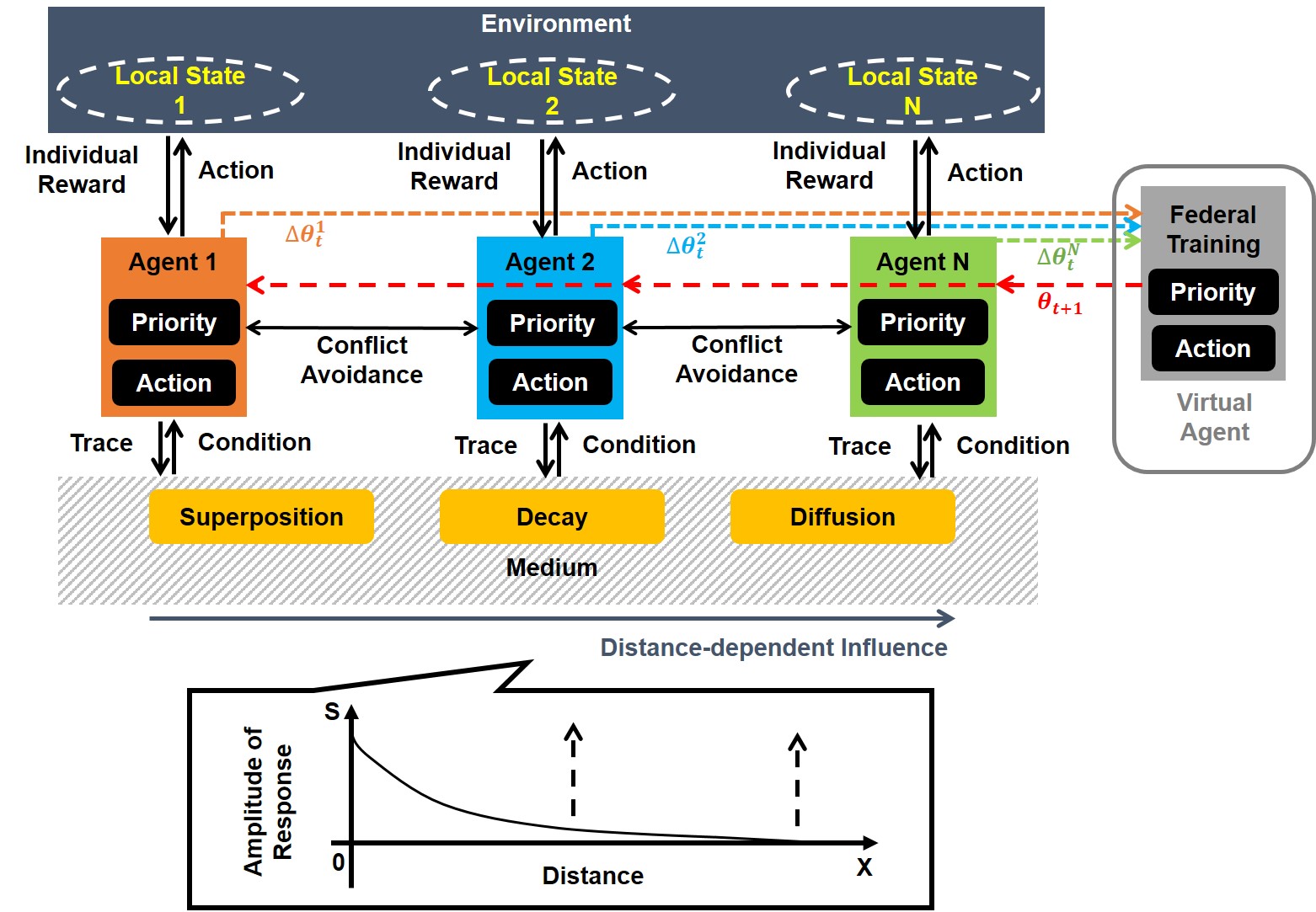}	
	\caption{Framework of stigmergic independent reinforcement learning (SIRL).}
	\label{fig1}
\end{figure*}

\section{Related Work}
A prototype of the stigmergy mechanism can be widely observed in natural colonies. Pagán \emph{et al.} \cite{On2019The} described a colony of social insects as a super-organism with brain-like cognitive abilities. This super-organism consists of a large number of small insect brains coupled with appropriate cooperative mechanisms. Despite its limited size, the small brain of each insect is capable of performing an adaptive learning process, which is similar to RL \cite{Holland1996,Arulkumaran2017Deep}. As a classical mechanism explaining the cooperative behavior of social insects \cite{Dorigo2000Ant}, stigmergy also includes a large number of small-scale learning processes \cite{Ricci2006Cognitive,Yong2009CoevolutionOR} and records their effect on the surrounding environment with the involved medium.

The application of stigmergy has been studied in the field of MAS \cite{Schoonderwoerd1996Ant,Caro1999AntNet,Dorigo2000Ant,Maru2017QoE,Takahashi2012A,Kanamori2013Evaluation}. Stigmergy is generally used to coordinate the behavior of multiple agents to accomplish the target task more efficiently. In particular, the coordination process in most applications focuses on the maintenance of digital pheromone, which is an important part of stigmergy, while the involved agent itself lacks the ability to learn the behavioral policy. For example, the coordination process in the classical ant colony optimization (ACO) algorithm \cite{Dorigo2007Ant} leads to an increased concentration of the correct pheromone; however, the behavioral policy of the involved agent is predetermined, that is, choosing among several concentrations in a probabilistic manner. In practice, this is effective in a bottom-to-up designed MAS \cite{2008Comparative} in which the behavioral policy of the involved agent can generally be predetermined and it is easier to predict the system's final performance. However, in many practical scenarios, the behavioral policies of the involved agents cannot be predetermined, and the agents must adjust their own policies while maintaining coordination.

In MARL, each agent can learn its behavioral policy through interaction with the surrounding environment. However, MARL also faces several challenges, such as the partial observation problem and credit assignment problem \cite{Busoniu2008A,Littman1994Markov}. Several studies have attempted to combine stigmergy with MARL. Aras \emph{et al.} \cite{ArasStigmergy} conceptually described how certain aspects of stigmergy can be imported into MARL, and defined an inter-agent communication framework. Verbeeck \emph{et al.} \cite{VerbeeckStigmergy} investigated the impact of an ant algorithm (AA) on the extension of RL to an MAS, and explained that stigmergy is essential in systems in which agents do not fully cooperate. However, the above studies all lack available algorithms. For an agent to perform well in a partially observable environment, it is usually necessary to require its actions to depend on the history of observations \cite{ShayeganDeep}. Peshkin \emph{et al.} \cite{Peshkin99learningpolicies} explored a stigmergic approach in which an external memory is created and included as part of the input to the involved agent. In addition, to optimize packet scheduling in routers, Bournenane \emph{et al.} \cite{2007Reinforcement} presented a pheromone-Q learning approach that combines the Q-learning algorithm \cite{1992Q} with a synthetic pheromone that acts as a communication medium. Specifically, the synthetic pheromone is introduced into the Q-learning updating equation through a belief factor. However, the effectiveness of these methods should be verified in a more complex MAS. Verbeeck \emph{et al.} \cite{Verbeeck2002Colonies} first proposed using learning automaton to design and test stigmergic communication. Furthermore, Masoumi \emph{et al.} \cite{2011Speeding} proposed to use the concept of stigmergy to calculate the reward received by each learning automaton (i.e., agent) to accelerate the learning process in Markov games. Similarly, Xu \emph{et al.} \cite{Xing2019Brain} proposed to use the digital pheromone to coordinate the behavior of multiple agents attempting to form a target shape. In particular, the distribution of digital pheromones helps provide guidance for the movement of agents. However, none of the above-mentioned works incorporate the advantageous capability of RL into the involved agents' decision-making processes. In contrast to the above-mentioned studies, our work focuses on introducing DRL into the decision-making process of each involved agent. 

Stochastic gradient descent based distributed machine learning algorithms have been studied in the literature in terms of both theoretical convergence analysis \cite{ShamirStochastic} and real-world experiments \cite{Dean2013Large}. Moreover, in addition to the traditional distributed algorithms, Rui \emph{et al.} \cite{Rui2020CPFed} proposed to use the federated learning algorithm to reduce the communication cost to train one neural network. Federated learning distributes the training task (e.g., face recognition) among several devices and obtains a common neural network through the integration process. Similarly, Sartoretti \emph{et al.} \cite{2018Distributed} proposed a distributed RL algorithm for decentralized multi-agent collection. This algorithm allows multiple agents to learn a homogeneous, distributed policy, where various agents work together toward a common target without explicit interaction. Sartoretti \emph{et al.} \cite{2018Distributed} also demonstrated that the aggregation of experience from all agents can be leveraged to quickly obtain a collaborative behavioral policy that naturally scales to smaller and larger swarms. However, the above-mentioned distributed learning algorithms usually prohibit interactions among agents to stabilize the learning process, which may be harmful, especially in cases where agents do not fully cooperate. Therefore, in this study, the stigmergy mechanism is specifically incorporated into the proposed distributed learning algorithm to achieve effective collaboration between agents while allowing each agent to retain efficient RL ability.

\vspace{0.9cm}
\section{System Framework}
We present the framework of the SIRL mechanism in Fig. \ref{fig1}. In particular, each agent is designed to learn independently during the training phase, and is required to act automatically during the decentralized execution phase. Note that each agent can only observe the environment partially and locally. Therefore, as illustrated at the bottom of Fig. \ref{fig1}, we deploy stigmergy as an indirect communication bridge between independent learning agents, which represents an explicit feedback loop between agents and the medium. Furthermore, as illustrated in the center of Fig. 1, a conflict-avoidance mechanism is deployed among adjacent agents to further reduce their behavioral localities. At the top of Fig. \ref{fig1}, we introduce the federal training method by appending a virtual agent to effectively optimize the neural network of each agent. To more easily clarify the framework and process of the SIRL mechanism, we primarily consider the flight formation of UAVs to monitor a specific target area as a typical example. We assume that multiple UAVs (i.e., agents) are flying and collaborating to form a specific team shape to hover above the final target area, and that each UAV must determine its flying policy independently based on its limited local environmental information, such as its relative position to other UAVs. In this scenario, the proposed framework can be applied to improve the learning effectiveness in terms of the final similarity between the target team shape and the shape formed by the UAVs or the aggregated cost of the UAVs.

\subsection{Stigmergy Mechanism}
In general, stigmergy consists of four main components (i.e., medium, trace, condition, and action), which together form a feedback loop between agents and their surrounding environment \cite{Heylighen2016Stigmergy1}, as illustrated in Fig. \ref{fig1}. Note that the medium can also be regarded as part of the entire environment. Here, the environment and medium are represented separately by different parts in Fig. \ref{fig1} to distinguish the traditional learning environment in RL from that utilized in the stigmergy mechanism. In addition, a trace (i.e., digital pheromone) is normally left by an agent in the medium as an indicator of the environmental change resulting from its action. Several traces left by different agents in the medium can diffuse and further mix in a spontaneous manner \cite{Dorigo2000Ant}. Then, the variation pattern of these digital pheromone traces is returned as the inter-influence to other agents for their subsequent actions, while the amplitude of this inter-influence is largely related to the inter-distance between agents \cite{Xing2019Brain}.

In addition, stigmergy can also serve as a potential solution to the decomposition of a global objective. For instance, the construction of a termite nest requires the cooperation of the entire colony, which usually passes through several generations. However, a single termite in this colony is unaware of the global objective (i.e., building a termite nest), due to the limited size of its brain. The cooperative mechanism utilized by this colony must be able to decompose the global objective into several small tasks that can be perceived by a single termite. Therefore, in addition to realize the indirect communication, stigmergy can also achieve decomposition of the global objective implicitly to help obtain individual reward in multi-agent collaboration.

\subsection{Conflict-Avoidance Mechanism}
\begin{figure}
	\centering
	\includegraphics[width=0.4\textwidth]{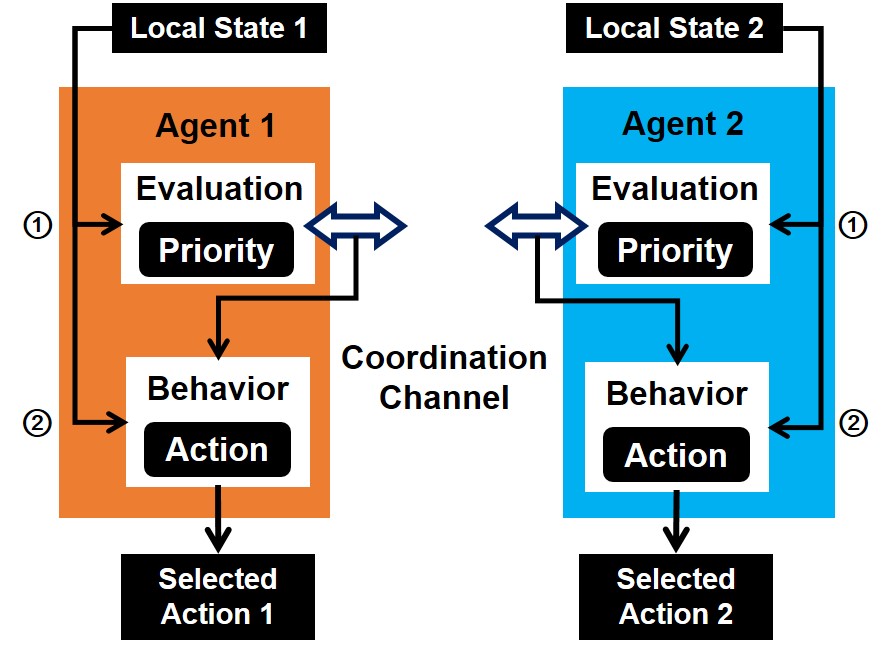}	
	\caption{Intuitive schematic of the conflict-avoidance mechanism.}
	\label{fig2}
\end{figure}

Conflicts between the actions of different agents may arise from the competition for limited task resources during multi-agent collaboration. To reduce the number of conflicts and minimize the amount of data that agents need to transmit during the collaboration process at the same time, we propose a conflict-avoidance mechanism by calculating action priorities for different agents. As illustrated in Fig. \ref{fig2}, Agent 1 (or 2) represents an independent learning agent in Fig. \ref{fig1}. In particular, there are two different internal neural network modules in each agent: the Evaluation Module and Behavior Module. The Evaluation Module is used to efficiently calculate the action priority of an agent at the current local state, which is further used to compete for the action opportunity. The Behavior Module is used to select appropriate actions for an agent according to the local input state when obtaining the action opportunity. Note that the action policy from the Behavior Module may be self-centered, signifying that each agent may be trapped in its local optimality while ignoring the global objective. Thus, the conflict-avoidance mechanism enabled by the Evaluation Module can help eliminate self-centered action policies to facilitate collaboration.

With the conflict-avoidance mechanism, each agent passes through two steps to determine the action at the current state. As illustrated in Fig. 2, in the first step, the input related to the current state of an agent is first sent to the Evaluation Module to calculate the action priority. The action priority value is then compared with that of nearby agents through a coordination channel, and a priority list within a small range can be obtained. In the second step, the same input can be sent to the Behavior Module for selecting appropriate actions only when an agent has the highest action priority in the first step. Otherwise, the agent remains silent.

\subsection{Federal Training Method}
Despite its success in the training of deep Q-learning networks \cite{Volodymyr2015Human,Schaul2015Prioritized}, experience replay may not be as effective in an MAS. Since a single agent in an MAS faces different tasks or situations, samples stored in the experience pool may not adapt to these changes. Recently, an asynchronous method for the advantage actor-critic (A2C) algorithm has been proposed and its advantages have been verified through training human-level players for Atari games \cite{Grondman2012A,Mnih2016Asynchronous,Bellemare2013The}. Here, we further propose a federal training method by improving the asynchronous A2C (i.e., A3C) algorithm to synchronously optimize the neural network of each agent through the average optimization of neural network gradients.

In this federal training method, each agent attempts to optimize its neural network not only through self-related experience, but also through neural network gradients from other collaborative teammates. Suppose that the number of active agents participating in the collaboration at time step $t$ is $N_t$, and $N_t \le N$, where $N$ denotes the total number of agents. Moreover, $N_t$ can also represent the number of agents that have obtained action opportunities through the conflict-avoidance mechanism. Gradients from these participating agents naturally form a mini-batch whose functionality is similar to that in the experience replay, and may be even more uncorrelated because they are sampled from different situations. Therefore, as illustrated in the right part of Fig. \ref{fig1}, a virtual agent is designed and added into SIRL, aiming to collect various local gradients of the involved agents for the average optimization. This virtual agent has the same neural network structure as other agents, but takes no action. 

\vspace{0.5cm}
\section{SIRL Algorithms for Multi-Agent Collaboration}
In this section, we provide additional details and mathematical formulations regarding the three above-mentioned mechanisms. We assume that $N$ agents are located in an area space and are dedicated to collaboratively fulfilling a specific task (e.g., UAVs to form a particular position shape). As illustrated in Figs. 1 and 2, at each time step $t$, each agent $i$ receives a local state $s_t^{(i)}$ from the environment's local state space $\mathcal{S}$. Depending on the local state, an action $a_t^{(i)}$ is selected from the individual action set $\mathcal{A}$ by each of $N_t$ agents. After the selected action is performed, an individual reward $r_t^{(i)}$ is returned to each participating agent to calculate the neural network gradients according to the loss function, $Loss(\theta_t^{(i)})$, so as to adjust its neural network parameters $\theta_t^{(i)}$. Specifically, the local state received by each agent is impacted by some particular environmental attractors, which are selected according to the digital pheromone, as discussed in Section \uppercase\expandafter{\romannumeral4}-A. Furthermore, because the selected action of each agent is influenced by its priority at the current local state, we discuss the related calculation method of this action priority in Section \uppercase\expandafter{\romannumeral4}-B. Finally, we present the entire training process of the two neural network modules of each agent using the federal training method. The main notations used in this paper are listed in TABLE I.

\begin{table}
	\centering
	\caption{Main Notations Used in This Paper.}
	\label{tb1}
	\begin{tabular}{ll}
		\toprule[1.1pt]
		Notation                   & Definition     \\
		\toprule[1.1pt]
		$N$    & Number of agents           \\
		$N_t$  & Number of active agents at time $t$\\
		$\mathcal{S}$    & Local state space\\
		$\mathcal{A}$    & Individual action set\\
		$s$    & Local state \\
		$a$    & Selected action\\
		$\pi$  & Action policy\\
		$r$    & Individual reward\\
		$R$    & Accumulated individual reward\\
		$\widetilde{r}$ & Deterministic individual reward\\
		$\widetilde{R}$ & Accumulated deterministic individual reward\\
		$l$    & Learning rate\\
		$e$    & State value network within Evaluation Module\\
		$\overline{e}$ & Target state value network within Evaluation Module\\
		$p$    & Policy network within Behavior Module\\
		$b$    & State value network within Behavior Module\\
		$\overline{b}$ & Target state value network within Behavior Module\\
		$V$    & State value\\
		$\theta$ & Neural network parameters\\
		$d$    & Distance between agent and attractor\\
		$D$    & Distance-dependent function\\
		$\varepsilon$ & Amount of digital pheromone\\
		$\xi$  & Attractor set\\
		$r_m$  & Individual reward from medium\\
		$v$    & Virtual agent\\		    
		$SI$   & Similarity\\
		\bottomrule[1.1pt]
	\end{tabular}
\end{table}

\subsection{Attractor Selection within the Stigmergic State Space}
The effectiveness of stigmergy can be improved by utilizing the digital pheromone \cite{Parunak2004Digital}. Unlike chemical pheromones left by ants in natural colonies, digital pheromones generated by intelligent agents are virtual and can be represented by several informative records in memory with attributes such as value, time, and location \cite{Mamei2005}. Furthermore, during swarm foraging, most ants are attracted to these chemical signals, whose distribution naturally forms a pheromone map between the food destination and nest. Similarly, a digital pheromone map that contains the distribution of digital pheromones for providing relevant information of the state space in the entire activity area is also deployed in SIRL. The entire digital pheromone map can be stored in a centralized manner in the virtual agent, or can be split into several parts and stored in a decentralized manner in several specified agents \cite{Parunak2004Digital}. Moreover, the digital pheromone map is continuously updated by mutual communication between its maintainer (e.g., virtual agent) and other agents in the activity area.

In SIRL, the digital pheromone is regarded as the trace left by an agent in the medium, while the digital pheromone map is regarded as the corresponding medium. As indicated by the dynamic features of the medium in Fig. \ref{fig1}, the digital pheromone experiences different evolution processes in the medium, which should be carefully designed to make the returned conditions more effective for each agent. Inspired by phenomena in natural colonies, in which chemical pheromones left by different ants can be superposed to increase the total influence, we model the accumulation of digital pheromones with different sources as a linear superposition. Moreover, instead of being restricted to a single area, the digital pheromone with larger amount will diffuse into surrounding areas. Furthermore, the amount of digital pheromone will decay over time. Therefore, the maintainer of the digital pheromone map should follow the three key principles: 1) linearly superposing digital pheromones with different sources in the same area, 2) diffusing digital pheromones into surrounding areas at a small scale with a fixed diffusion rate after a new digital pheromone has been left, and 3) decreasing the amount of digital pheromone at positions already occupied by agents with a fixed decay rate. Note that the decay and diffusion rate are both constants between $0$ and $1$.

With a digital pheromone map for the stigmergic state space, each agent can sense the amount of digital pheromone within a certain range. Without loss of generality, we take the aforementioned UAV application scenario as an example. Here, we regard any block (i.e., unit area) filled with the digital pheromone as an attractor in the local environment, which has an attractive effect on any nearby mobile agent for efficiently observing the local state space. Similar to the classical ACO algorithm, within the local state space, each intelligent agent can independently perform its action (i.e., approaching an attractor) by selecting a suitable attractor within its sensing range from several potential attractor 
candidates, which can be expressed by

\begin{equation}
C_{i,j}(t)=\frac{D(d_{i,j}(t))\cdot \varepsilon_j(t)}{\sum_{j\in\xi_i(t)}D(d_{i,j}(t))\cdot \varepsilon_j(t)},
\end{equation}
where $C_{i,j}(t)$ is the probability of agent $i$ selecting attractor $j$; $\varepsilon_j(t)$ is the total amount of digital pheromone in attractor $j$ at time $t$; $\xi_i(t)$ is the set of attractors within the sensing range of agent $i$; $d_{i,j}(t)$ is the Euclidean distance between agent $i$ and attractor $j$; and $D(\cdot)$ is a monotone function used to reduce the effect of the digital pheromone as the inter-distance $d_{i,j}(t)$ increases \cite{Xing2019Brain}, which is intuitively illustrated at the bottom of Fig. \ref{fig1}. This attractor selection method is inspired by the classical ACO algorithm \cite{Dorigo2007Ant}. However, we use function $D(\cdot)$ to replace the heuristic factor, which is commonly used to consider the constrains of an optimization problem, of the original ACO algorithm to improve the selection process. Function $D(\cdot)$ can cause an agent to pay more attention to nearby influential attractors while avoiding the so-called ping-pong effect in the local environment. In addition, selecting attractors in a stochastic manner can cause agents to perform actions (i.e., approaching target positions) with a smaller amount of digital pheromone, and can prevent a large number of agents from crowding in a small local environment. 

The location of the selected attractor serves as an informative part of the input local state for each agent. Depending on the two neural network modules of each agent, an action is selected according to the input local state. Furthermore, following the stigmergic principle, any agent, which has performed the selected action accordingly leaves an additional digital pheromone in the medium, to provide new condition information for the subsequent selection of attractors. This process can be expressed by

\begin{equation}
\small
\varepsilon_j(t+1) = \left\{
\begin{array}{lcl}
{\varepsilon_j(t) + a_1}, &\text{if\ $\varepsilon_j$\ is\ in the labeled area};\\
{\varepsilon_j(t) \cdot b_1}, &\text{otherwise},   
\end{array}  
\right.
\end{equation}
where $a_1$ represents the fixed amount of digital pheromone left by an agent at a time, and $b_1$ is a discount constant between 0 and 1 that helps gradually remove useless attractors. The labeled area indicates that the agent has partially fulfilled a small task.

\subsection{Action Priority Determination}
In this subsection, we discuss the methods to calculate the action priority for the conflict-avoidance mechanism. Corresponding to the two neural network modules in Fig. 2, we exploit two algorithms to optimize their parameters, respectively. First, for the Evaluation Module, we define an internal state value network whose output is the expected accumulated deterministic individual reward at $s_{t}^{(i)}$:

\begin{equation}
V_e(s_{t}^{(i)};\theta_{e}^{(i)}) = \mathbb{E}[\widetilde{R}_{t}^{(i)}|s_{t}^{(i)} = s, a_{t}^{(i)} = (a;\theta_{p}^{(i)})],
\end{equation}
\begin{equation}\nonumber
a_t^{(i)} = \arg\max_{a\in \mathcal{A}}  \pi(s_{t}^{(i)},a;\theta_{p}^{(i)}),
\end{equation}
where $s_{t}^{(i)}$ represents the local state observed by agent $i$ at time $t$. Subscript $e$ represents the state value network in the Evaluation Module, and $\theta_{e}^{(i)}$ denotes the related parameters. Moreover, $V_e$ is the state value of $s_{t}^{(i)}$, which is regarded as the action priority of agent $i$ at the current local state. $a_{t}^{(i)}$ denotes the selected action toward the chosen attractor of agent $i$ at time $t$. Furthermore, subscript $p$ represents the policy network in the Behavior Module, while $\pi$ represents its action policy. Note that the evaluation of the action priority at $s_{t}^{(i)}$ is based on the deterministically executed action at the same local state, and the returned individual reward during the training process of the state value network in the Evaluation Module is also deterministic. Therefore, we define $\widetilde{R}_{t}^{(i)}$ as the accumulated deterministic individual reward of agent $i$ at time $t$, which is calculated by

\begin{equation}
\small
\widetilde{R}_{t}^{(i)} = \left\{
\begin{array}{lcl}
{\widetilde{r}_{t}^{(i)} + \gamma_2 \cdot V_{\overline{e}} (s_{t+1}^{(i)};\theta_{\overline{e}}^{(i)})}, &\text{if\ $s_{t+1}^{(i)}$\ is\ non-terminal};\\
{\widetilde{r}_{t}^{(i)}}, &\text{otherwise},   
\end{array}  
\right.
\end{equation}
where $\gamma_2$ is a discount factor, and $\widetilde{r}_{t}^{(i)}$ is the returned deterministic individual reward. Subscript $\overline{e}$ represents the target state value network of agent $i$, whose parameters and output are represented by 
$\theta_{\overline{e}}^{(i)}$ and $V_{\overline{e}}$, respectively. The target state value network is used to calculate the state value of the new input local state $s_{t+1}^{(i)}$ and further help calculate the accumulated deterministic individual reward, as illustrated in the first line of (4). Moreover, the target state value network is almost the same as the state value network in the Evaluation Module except that its parameters are periodically copied from the original state value network \cite{Volodymyr2015Human}.
Finally, the loss function of the state value network in the Evaluation Module can be expressed as

\begin{equation}
Loss(\theta_{e}^{(i)}) = 0.5 \cdot [\widetilde{R}_{t}^{(i)} - V_e(s_{t}^{(i)};\theta_{e}^{(i)})]^{2}.
\end{equation} 

For the Behavior Module, we use the A2C algorithm to optimize its parameters \cite{Grondman2012A}. In particular, there is a policy and a state value network in the Behavior Module that share the same input local state from the local environment, including the stigmergic state space with attractor indexing. Their loss functions, which are used to calculate the gradients of the neural network parameters, are respectively expressed as

\begin{equation}
Loss(\theta_{p}^{(i)}) = - \mathrm{log} \pi(a_{t}^{(i)}|s_{t}^{(i)};\theta_{p}^{(i)})(R_{t}^{(i)} - V_{b}(s_{t}^{(i)};\theta_{b}^{(i)})),
\end{equation}

\begin{equation}
Loss(\theta_{b}^{(i)}) = 0.5 \cdot [R_{t}^{(i)} - V_{b}(s_{t}^{(i)};\theta_{b}^{(i)})]^2,
\end{equation}
where subscripts $p$ and $b$ represent the policy and state value network in the Behavior Module, respectively. The local state $s_{t}^{(i)}$ is sent to two different neural networks (i.e., the policy network and the state value network) to calculate the corresponding output. For the policy network, the output is the action policy $\pi$. We select each action in a probabilistic manner during the training phase of the policy network; thus, the returned individual reward is stochastic. For the state value network, the output is the state value of $s_{t}^{(i)}$ (i.e., $V_{b}$), which is used to calculate the advantage (i.e., $R_{t}^{(i)} - V_{b}(s_{t}^{(i)};\theta_{b}^{(i)})$) to speed up the convergence of the action policy in the parallel policy network. Accordingly, $R_{t}^{(i)}$ is the accumulated individual reward of agent $i$, which can be expressed as

\begin{equation}
\small
R_{t}^{(i)} = \left\{
\begin{array}{lcl}
{r_{t}^{(i)} + \gamma_1 \cdot V_{\overline{b}}(s_{t+1}^{(i)};\theta_{\overline{b}}^{(i)})}, &\text{if\ $s_{t+1}^{(i)}$\ is\ non-terminal};\\
{r_{t}^{(i)}}, &\text{otherwise},   
\end{array}  
\right.
\end{equation}
where $r_{t}^{(i)}$ is the individual reward received by agent $i$ at time $t$, and $\gamma_1$ is a discount factor that can normally be set to a constant between 0 and 1 (e.g., 0.9). Similarly, subscript $\overline{b}$ represents another target state value network of agent $i$. Note that under the conflict-avoidance mechanism, any agent with the highest action priority will have the opportunity to perform the selected action. Because the execution order of different actions is arranged by their accumulated deterministic individual rewards, this estimation method of action priority is expected to obtain a large global reward.

\subsection{Training Neural Network Modules}
\begin{figure}
	\centering
	\includegraphics[width=0.45\textwidth]{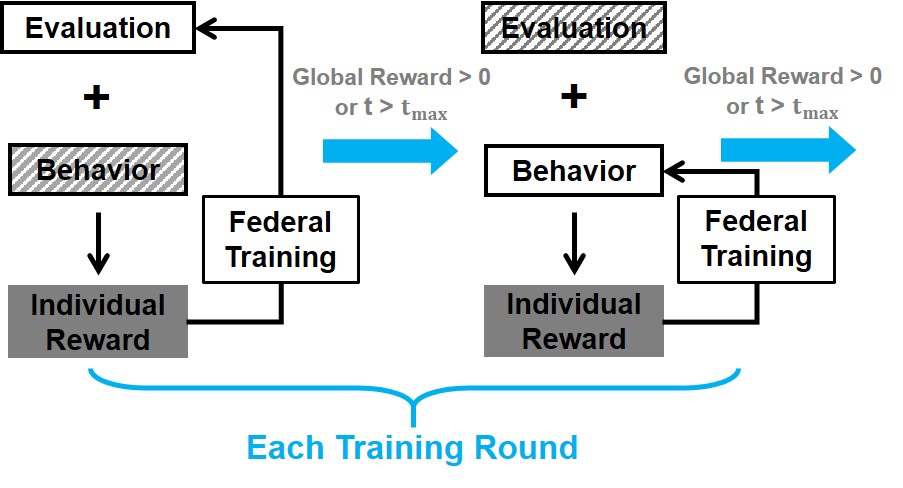}	
	\caption{Two successive sessions in each training round.}
	\label{fig3}
\end{figure}

Under the conflict-avoidance mechanism, the Evaluation and Behavior Modules work together to obtain an individual reward, and forge a cooperative relationship. Furthermore, according to (3), the estimation of the action priority is also based on the deterministic action policy of the Behavior Module. Therefore, there are two successive sessions in each training round (i.e., training epoch). In the left part of Fig. \ref{fig3}, we freeze the parameters of the Behavior Module in the training session of the Evaluation Module. Similarly, we freeze the parameters of the Evaluation Module in the training session of the Behavior Module, which is indicated in the right part of Fig. \ref{fig3}. The federal training method is applied in both sessions, where a virtual agent is also employed. Furthermore, we set a terminal condition for both training sessions. Each session is stopped as the number of internal time steps reaches its maximum, or the global reward is positive. The performance of multi-agent collaboration is represented by the global reward, whose improvement can implicitly indicate the convergence of the current training of the neural network modules. In Fig. 3, $t$ denotes the index of time steps while $t_{\mathrm{max}}$ represents its maximum. $t$ is set to $0$ at the beginning of a training session. During each training round, a sample is sent to both sessions to optimize different neural network modules. 

In MARL, the global reward can typically be used to indicate the performance of multi-agent collaboration. However, it usually cannot be directly used as the individual reward received by an agent, as the global reward is determined by a series of actions from different agents. In general, the individual reward can be obtained through a reasonable decomposition of the global objective. In SIRL, after the selected action is performed, each agent receives an individual reward from the medium. Because the objective of each agent is to approach its selected attractor in the stigmergic state space, we define the returned individual reward as the Euclidean measure (i.e., inter-distance change) between the position of each agent and its selected attractor, which can be expressed as

\begin{equation}
r_m^{(i)}(t) = \rho_1 \cdot \text{max}\left([d_{i,j}(t-1) - d_{i,j}(t)],0\right),
\end{equation} 
where the subscript $m$ represents the medium in SIRL, and $r_m^{(i)}(t)$ represents the individual reward received by agent $i$ from the medium at time $t$. In addition, $\rho_1$ is a scalar factor, while $d_{i,j}(t)$ represents the inter-distance between agent $i$ and its selected attractor $j$, where $j \in \xi_i(t-1)$. Note that the reward $r_m^{(i)}(t)$ is obtained due to the implementation of the digital pheromone, which indicates the decomposition process of the stigmergy mechanism for the global objective. In particular, during each training round, $r_m^{(i)}(t)$ is set to $\widetilde{r}_{t}^{(i)}$ in (4) during the training session of  the Evaluation Module, and $r_{t}^{(i)}$ in (8) during the training session of the Behavior Module.  

Furthermore, as illustrated in Figs. 1 and 3, the federal training method is used to optimize the parameters of different neural network modules through the average optimization: 

\begin{equation}
\theta_{t+1}^{(v)} = \theta_{t}^{(v)} + v_{t}^{(v)},
\end{equation}
\begin{equation}
v_{t}^{(v)} = \rho \cdot v_{t-1}^{(v)} - l_t \cdot \frac{1}{N_{t}} \sum_{i=1}^{N_{t}} \frac{\partial Loss(\theta_{t}^{(i)})}{\partial \theta_{t}^{(i)}},
\end{equation}
where the superscript $v$ represents the virtual agent. At time step $t=0$, $v_{0}^{(v)}$ is set to $0$. $\rho$ is a momentum factor, while $l_t$ denotes the learning rate of parameters. The federal training method is inspired by the A3C algorithm; however, it applies a synchronous updating method, which has been demonstrated to have a lower error bound \cite{DuttaSlow}. Moreover, we add a momentum factor in the updating process to accelerate the convergence. Because a virtual agent has the same neural network structure as other agents, it can be used to optimize the parameters of different neural network modules. For simplicity, we use $\theta_{t}^{(i)}$ in (11) to represent the parameters of either the Evaluation or Behavior Module for agent $i$, and use $\theta_{t}^{(v)}$ to represent the parameters of the same neural network module of the virtual agent. In particular, for the current training of the neural network module, gradients of the involved neural network parameters are first calculated in these $N_t$ agents according to the corresponding loss function, $Loss(\theta_{t}^{(i)})$, and are then sent to the virtual agent for the average optimization. Finally, the newly calculated parameters $\theta_{t+1}^{(v)}$ are sent back to all agents to update their neural network modules. The entire SIRL algorithm can be divided into a training and testing part, which are detailed in Algorithms 1 and 2, respectively.
		
\begin{algorithm*}
	\footnotesize
	\caption{\textbf{The training part of SIRL.}}
	\begin{algorithmic}[1]
		\REQUIRE Agents with the neural network modules, the sample set, the target shape, number of training rounds;
		\ENSURE Agents with the well-trained neural network modules;
		\STATE \textbf{Initialize} the whole activity area, the digital pheromone map, the labeled area, the neural network modules within each agent, diffusion rate, decay rate, $t_{\mathrm{max}}$, time step $t$, the range of sensing digital pheromone, the range of coordination channel;
		\FOR {each training round}
		\STATE $\textbf{//* Training\ session\ of\ Evaluation\ Module *//}$
		\STATE Select a sample randomly from the sample set and initialize the location of agents according to the selected sample;
		\WHILE {$t \le t_{\mathrm{max}}$} 
		\FOR {each agent}
		\STATE Select an attractor according to (1) and form the input local state;
		\STATE Send the local state to Evaluation Module;
		\STATE Send the same local state to Behavior Module and select an action with the largest probability;
		\STATE Perform the action;
		\STATE Modify the digital pheromone at current position according to (2);
		\IF {the extra digital pheromone is left}
		\STATE Diffuse the digital pheromone to the surrounding areas with the fixed diffusion rate;
		\STATE Superpose the amount of digital pheromone at the same position linearly;
		\ENDIF
		\STATE Calculate the individual reward according to (9);
		\STATE Select a new attractor according to (1) and form the new input local state;
		\ENDFOR
		\STATE Decay the amount of digital pheromone at positions already occupied by agents with the fixed decay rate;
		\IF {the calculated global reward $>$ 0}
		\STATE Break;
		\ELSE
		\FOR {each agent}
		\STATE Calculate the gradients $\frac{\partial Loss(\theta_{e}^{(i)})}{\partial \theta_{e}^{(i)}}$ of the state value network within Evaluation Module according to (4) - (5) with the new input local state;
		\STATE Send the calculated gradients to the virtual agent;
		\ENDFOR
		\STATE The virtual agent receives the gradients from agents and optimizes the internal state value network within Evaluation Module according to (10) - (11);
		\STATE The virtual agent sends back the calculated parameters $\theta_{t+1}^{(v)}$ to all agents;
		\STATE Each agent updates the state value network within Evaluation Module with $\theta_{t+1}^{(v)}$;
		\ENDIF
		\ENDWHILE
		\STATE $\textbf{//* Training\ session\ of\ Behavior\ Module *//}$
		\STATE Initialize the location of agents according to the selected sample;
		\WHILE {$t \le t_{\mathrm{max}}$}
		\FOR {each agent}
		\STATE Select an attractor according to (1) and form the input local state;
		\STATE Send the local state to Evaluation Module and calculate the action priority;
		\STATE Send out the action priority through the coordination channel and receive the returned priority list;
		\IF {the own action priority is the largest}
		\STATE Send the same local state to Behavior Module and select an action in a probabilistic manner;
		\STATE Perform the action;
		\STATE Modify the digital pheromone at current position according to (2);
		\IF {the extra digital pheromone is left}
		\STATE Diffuse the digital pheromone to the surrounding areas with the fixed diffusion rate;
		\STATE Superpose the amount of digital pheromone at the same position linearly;
		\ENDIF
		\STATE Calculate the individual reward according to (9);
		\STATE Select a new attractor according to (1) and form the new input local state;
		\ENDIF
		\ENDFOR
		\STATE Decay the amount of digital pheromone at positions already occupied by agents with the fixed decay rate;
		\IF {the calculated global reward $>$ 0}
		\STATE Break;
		\ELSE
		\FOR {each agent getting the action opportunity}
		\STATE Calculate the gradients $\frac{\partial Loss(\theta_{p}^{(i)})}{\partial \theta_{p}^{(i)}}$  and $\frac{\partial Loss(\theta_{b}^{(i)})}{\partial \theta_{b}^{(i)}}$ of the policy and state value networks within Behavior Module according to (6) - (8) with the new input local state;
		\STATE Send the calculated gradients to the virtual agent;
		\ENDFOR
		\STATE The virtual agent receives the gradients from agents and optimizes the internal policy and state value networks within Behavior Module according to (10) - (11);
		\STATE The virtual agent sends back the calculated parameters $\theta_{t+1}^{(v)}$ to all agents;
		\STATE Each agent updates the policy and state value networks within Behavior Module with $\theta_{t+1}^{(v)}$;
		\ENDIF
		\ENDWHILE		
		\ENDFOR
		\STATE \textbf{Return} agents with the well-trained neural network modules;
	\end{algorithmic}
\end{algorithm*}

\begin{algorithm*}
	\footnotesize
	\caption{\textbf{The testing part of SIRL.}}
	\begin{algorithmic}[1]
		\REQUIRE Agents with the well-trained neural network modules, the location of each agent, the target shape, number of iterations;
		\ENSURE The final similarity;
		\STATE \textbf{Initialize} the whole activity area, the digital pheromone map, the labeled area, diffusion rate, decay rate, time step $t$, the range of sensing digital pheromone, the range of coordination channel;
		\FOR {each iteration}
		\FOR {each agent}
		\STATE Select an attractor according to (1) and form the input local state;
		\STATE Send the local state to Evaluation Module and calculate the action priority;
		\STATE Send out the action priority through the coordination channel and receive the returned priority list;
		\IF {the own action priority is the largest}
		\STATE Send the same local state to Behavior Module and select an action with the largest probability;
		\STATE Perform the action;
		\STATE Modify the digital pheromone at current position according to (2);
		\IF {the extra digital pheromone is left}
		\STATE Diffuse the digital pheromone to the surrounding areas with the fixed diffusion rate;
		\STATE Superpose the amount of digital pheromone at the same position linearly;
		\ENDIF
		\ENDIF
		\ENDFOR
		\STATE Decay the amount of digital pheromone at positions already occupied by agents with the fixed decay rate;
		\ENDFOR
		\STATE Calculate the final similarity according to the target shape;
		\STATE \textbf{Return} the final similarity;
	\end{algorithmic}
\end{algorithm*}

\vspace{0.2cm}
\section{Experimental Simulation Settings and Numerical Results}
In this section, we create a simulation scenario in which a certain number of mobile agents (e.g., UAVs) in an activity area automatically form a specified team shape. In this simulation scenario, the target team shape is transferred through the binary normalization of an image that is taken from the standard Modified National Institute of Standards and Technology (MNIST) dataset \cite{Lecun1998Gradient}. The MNIST dataset is a large set of handwritten digits that is commonly used for training various image processing systems. In addition, it is widely used for training and testing in the field of machine learning. Specifically, each image in this dataset has a standard size (i.e., 28 $\times$ 28). As illustrated in Fig. 4, an agent is represented by a non-zero block (i.e., a pixel, as in digitized binary image). The white block (i.e., unit area) represents an agent, while black blocks represent the potential positions the mobile agent can visit. We set the distance between any two adjacent blocks to $1$. At the beginning, all mobile agents are distributed randomly across the entire activity area (i.e., the entire image). The experimental objective in this scenario can be mapped to many real-world multi-agent collaboration tasks, such as the flight formation of UAVs monitoring a certain target area. 

\begin{figure}
	\centering
	\subfigure[]{%
		\includegraphics[width=0.1\textwidth]{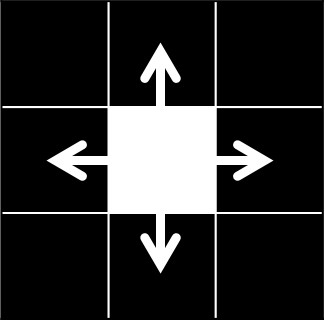}}%
	\hspace{.1in}
	\centering
	\subfigure[]{%
		\includegraphics[width=0.18\textwidth]{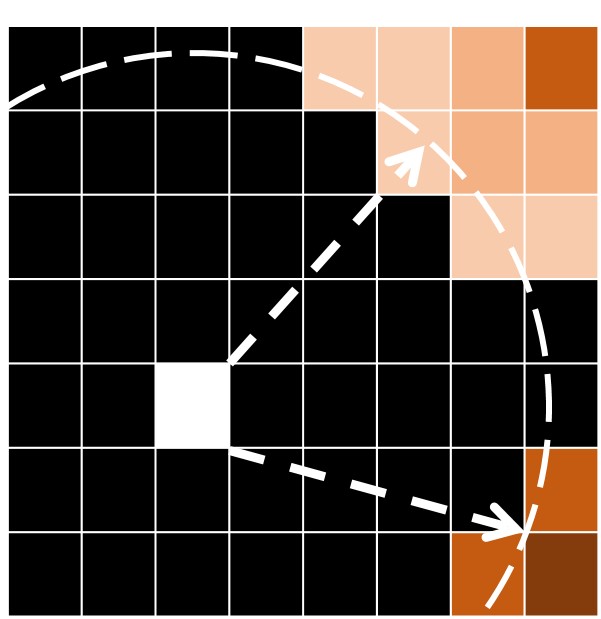}}%
	\caption{Moving mode and digital pheromone sensing range of a single mobile agent in the activity area.}
	\label{fig4}
\end{figure}

First, the simulation settings for the involved agents and the activity area are described as follows:

\begin{itemize}
	\item Each agent is supposed to move a block at each time step toward one of four directions: 1) Up, 2) Down, 3) Left, and 4) Right.
	\item Each agent can leave the digital pheromone in the current position and sense the amount of digital pheromones in the surrounding blocks within a certain range.
	\item Each agent can identify whether the occupied position is labeled or unlabeled.
	\item Each agent can sense the existence of neighbors in the Up, Down, Left, and Right directions to avoid collision.
	\item Each block in the activity area can only be occupied by one agent at a time.
	\item The activity area can be classified as labeled and unlabeled, where the former corresponds to the target team shape to be formed.
\end{itemize}

The moving mode of a single agent is illustrated in Fig. \ref{fig4}(a). In Fig. \ref{fig4}(b), the white dotted line represents the boundary of the digital pheromone sensing range of that agent. Brown blocks indicate the presence of digital pheromones, and colors with different shades represent distinct amounts of the pheromone. In addition, the coordination channel for the conflict-avoidance exists between any central agent and its eight possible neighbors, which are also called Moore neighbors in mobile cellular automation \cite{Miramontes1993Collective}. In the experimental simulation, we use a Gaussian function to play the role of $D(\cdot)$ in (1), which can be expressed as

\begin{equation}
D(d_{i,j}(t)) = a_2\cdot \exp(-\frac{(d_{i,j}(t) - b_2)^2}{2c_1^2}),
\end{equation}
where $a_2$ represents the peak value and is set to $1$; $b_2$ is the mean value and is set to $0$; and $c_1$ represents the standard deviation and is normally set to $0.25$.

\begin{figure*}
	\centering	
	\subfigure[]{%
		\includegraphics[width=0.4\textwidth]{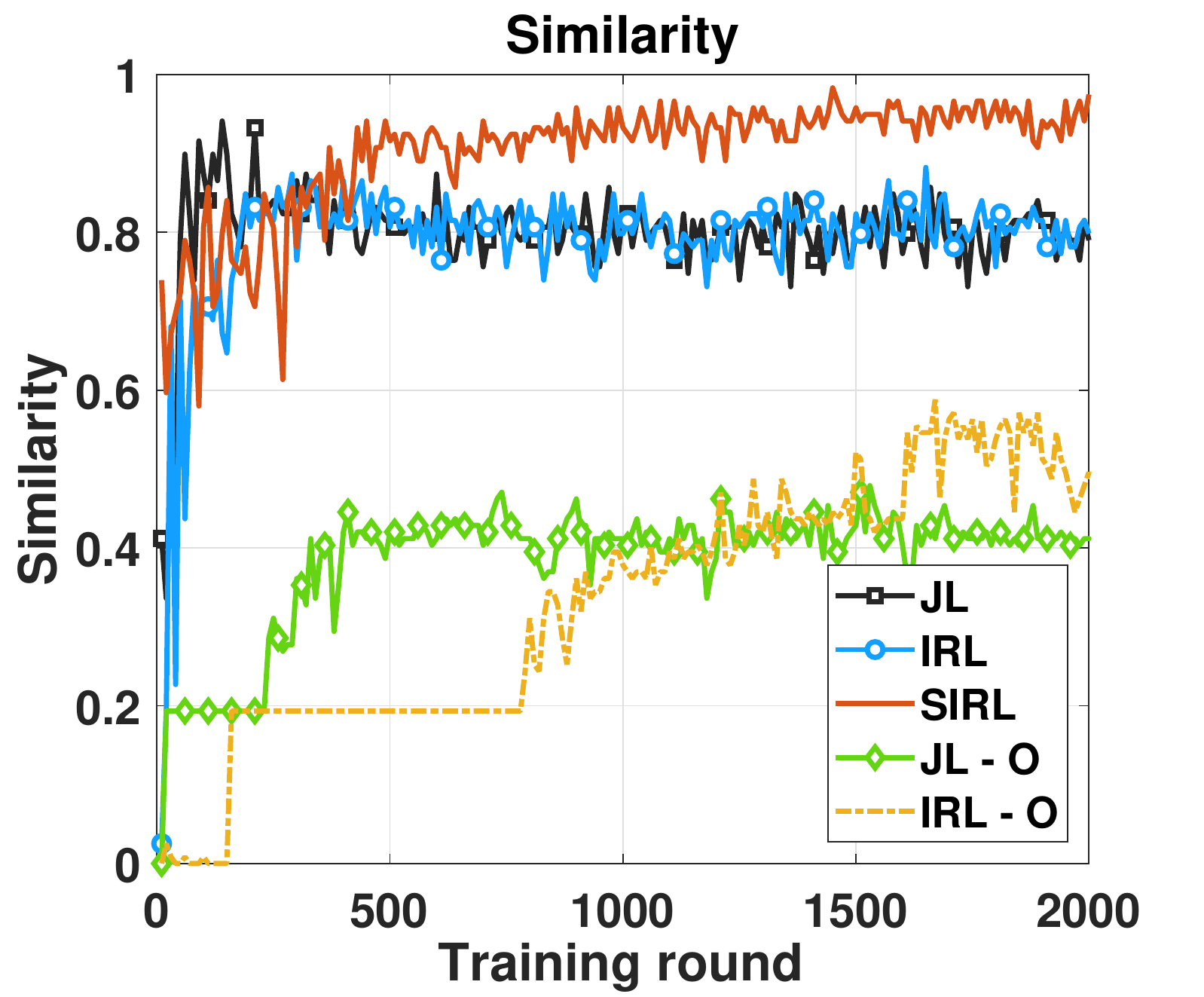}}%
	\subfigure[]{%
		\includegraphics[width=0.4\textwidth]{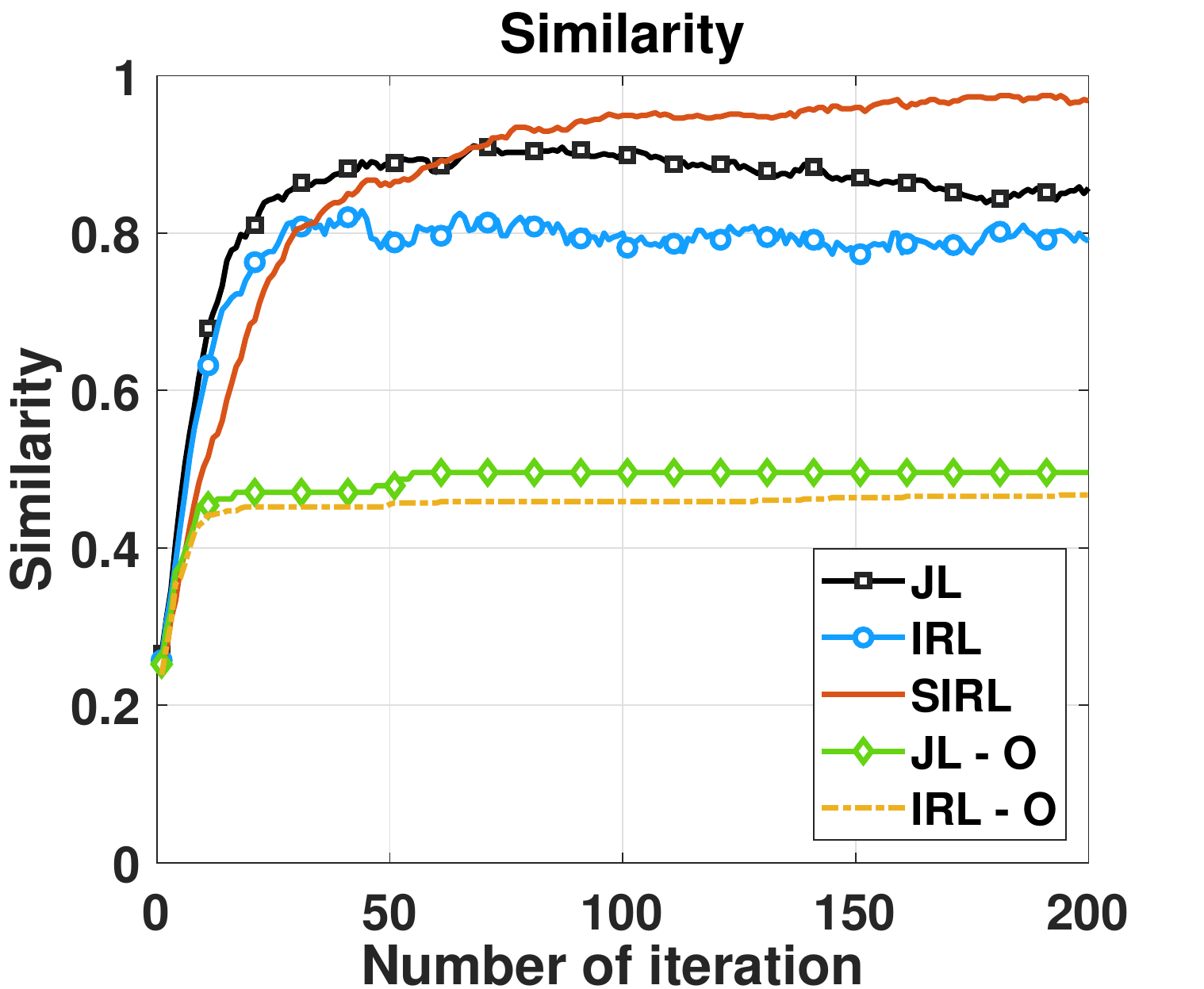}}%
	\caption{Training and testing performance of different methods in the task to form the shape ``4."}
	\label{fig5}
\end{figure*}

The similarity, $SI$, is calculated by the ratio of the number of agents that end up in the labeled area to the total number of agents in the activity area, which is further determined by the number of non-zero pixels required to form the target shape. In the training part, we define the increase in similarity after each time step (i.e., $\Delta SI$) as the global reward, which can be positive or negative. In addition, similar to the settings in \cite{Xing2019Brain}, we use the position distribution of all agents after a certain number of iterations as the swarm's initial state, which can also be regarded as a sample. Approximately 7,500 samples are extracted from different iterations in this simulation during the formation of the target shape ``4." In particular, the new position distribution of all agents is returned after each time step to calculate both the global and individual rewards so as to further optimize the neural network modules. Note that the neural network modules trained by this sample set can also be used in the testing process to form another shape (e.g., ``1," ``2," ``0," ``6," and ``8"). 

Furthermore, the Evaluation and Behavior Modules share the same input local state, which is denoted by a vector with seven elements. The first four elements represented by bit numbers are used to confirm whether there are neighbors in the following four adjacent positions: Up, Right, Down, and Left. The fifth and sixth elements are used to describe the relative position of the selected attractor in a two-dimensional plane, and the seventh element is used to confirm whether the current occupied position is labeled or unlabeled. The individual action set $\mathcal{A}$ contains five different actions: Up, Right, Down, Left, and Stop. However, since the local state received by each agent contains the relative positions of adjacent agents, the recorded new input local state $s_{t+1}^{(i)}$ may be inaccurate at time step $t+1$ due to the possible movement of adjacent agents at time step $t$, thus leading to inaccurate estimation of $V_{\overline{e}}(s_{t+1}^{(i)};\theta_{\overline{e}}^{(i)})$ in (4). In the following simulation, we use different values of $\gamma_2$ to test this phenomenon. 

\begin{figure}
	\centering
	\includegraphics[width=0.48\textwidth]{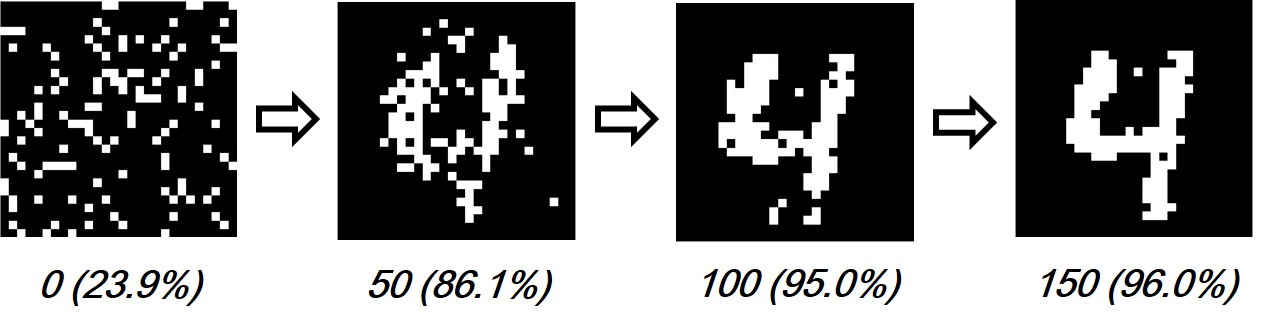}	
	\caption{Shapes formed at different iterations using stigmergic independent reinforcement learning (SIRL).}
	\label{fig6}
\end{figure}

\begin{table}
	\centering
	\caption{Received Individual Reward for Joint Learning-Origin (JL-O) and Independent Reinforcement Learning-Origin (IRL-O).}
	\label{tb2}
	\begin{tabular}{c|c}
		\toprule[1.1pt]
		Transition                                  & Reward \\
		\hline
		Unlabeled area to Unlabeled area: 0 $\to$ 0 & 0 \\
		Unlabeled area to Labeled area: 0 $\to$ 1  & $a_3$ \\
		Labeled area to Unlabeled area: 1 $\to$ 0   & 0 \\
		Labeled area to Labeled area: 1 $\to$ 1  & $b_3 \cdot$ max($\bigtriangleup SI$, 0)  \\
		\bottomrule[1.1pt]
	\end{tabular}
\end{table} 

We first present the convergence of SIRL and compare it with that of other typical methods. The first method, joint learning (JL), attempts to learn the optimal behavior from the joint information using only the Behavior Module, whose input is a cascaded vector containing all the input local states of surrounding agents within the Moore neighborhood, while the conflict-avoidance mechanism is disabled. The second method, IRL, has almost the same settings as JL except that its input vector only contains the self-related local state. The third method, joint learning-origin (JL-O), and fourth method, IRL-origin (IRL-O), are modified from the JL and IRL methods, respectively, by further disabling the stigmergy mechanism and replacing the attractors by the exact coordinates of the agents. In this situation, each agent receives a non-zero individual reward only when it enters the labeled area, and the global reward is also considered afterward. The received individual reward in this situation is indicated in TABLE \uppercase\expandafter{\romannumeral2}. Here, the transition 0 $\to$ 1 represents an agent moving from the unlabeled area to the labeled area. $a_3$ and $b_3$ are both positive constants.

The training and testing performance of the above-mentioned five methods is presented in Fig. \ref{fig5}, and an intuitive illustration of the formed shapes with respect to the iteration index in SIRL is presented in Fig. \ref{fig6}. In Fig. \ref{fig5}(a), the neural network modules are tested every 10 training rounds. It can be observed from Fig. \ref{fig5} that there is an evident performance difference between methods with and without the stigmergy mechanism, as indicated by the curves of SIRL, JL, and IRL, since stigmergy can better decompose the global objective and achieve a higher final similarity. In addition, although the joint information is obtained in JL, it is simply treated as a noisy signal by each agent without appropriate utilization. Therefore, despite different inputs, the performance of JL and IRL is almost identical. Moreover, SIRL performs better than JL or IRL,  benefiting from the conflict-avoidance mechanism, which can reconcile the contradictions caused by the actions of different agents and further improve the cooperation efficiency. 

We also present the training performance of SIRL in Fig. 7 when the discount factor $\gamma_2$ takes different values. We can observe that the training performance of SIRL declines as the value of $\gamma_2$ increases to $1$. In SIRL, based on the current local observations of the environment and the behavioral policy, the Evaluation Module of each agent is used to calculate the action priority among the action candidates. However, due to the existence and influence of other agents, the transition process from one state to another becomes unpredictable for each individual agent. In this situation, the reward mapped from the future state by the discount factor $\gamma_2$ becomes inaccurate. A smaller $\gamma_2$ can generally reduce this estimation error. Therefore, in the following simulation, we set the discount factor $\gamma_2 = 0$ in the training session of the Evaluation Module of each agent to limit the accumulation of $\widetilde{R}_t^{(i)}$ to only one step.

\begin{table*}
	\centering
	\caption{Final Similarities Using Different Methods.}
	\label{tb3}
	\begin{tabular}{c|c|c|c|c|c|c|c|c|c|c}
		\toprule[1.1pt]
		\multirow{2}{*}{Stigmergy} & \multirow{2}{*}{Conflict-Avoidance} & \multirow{2}{*}{Federal Training} & \multirow{2}{*}{Input Local State} & \multirow{2}{*}{Algorithm}  & \multicolumn{6}{c}{Shape (Number of involved agents)}\\
		\cline{6-11} ~&~&~&~&~&	1 (65)       & 4 (119)    & 2 (161)   & 0 (179)   & 6 (128)   & 8 (163)    \\
		\hline
		Applied                  & Applied & Applied & Self-related & SIRL          & $\bf{96.6\%}$ & $\bf{97.5\%}$ & $\bf{97.5\%}$ & $\bf{95.1\%}$ & $\bf{96.0\%}$ & $\bf{96.4\%}$ \\
		Applied                  & None    & Applied & Joint & JL            & $84.3\%$ & $85.7\%$ & $82.1\%$ & N/A        & N/A        & N/A        \\
		Applied                  & None    & Applied & Self-related & IRL           & $84.9\%$ & $83.2\%$ & $81.9\%$ & N/A        & N/A        & N/A        \\
		None                     & None    & Applied & Joint & JL - O        & $35.4\%$ & $49.6\%$ & $41.0\%$ & N/A        & N/A        & N/A        \\
		None                     & None    & Applied & Self-related & IRL - O       & $33.9\%$ & $46.7\%$ & $37.9\%$ & N/A        & N/A        & N/A        \\
		Applied                  & Applied & None    & Self-related & DC            & $99.7\%$ & $98.3\%$ & $94.5\%$ & $84.8\%$ & $88.4\%$ & $86.9\%$ \\
		Applied                  & Applied & None    & Self-related & CS            & $94.2\%$ & $95.1\%$ & $87.8\%$ & $57.2\%$ & $74.4\%$ & $75.3\%$ \\
		Applied                  & Applied & Applied & Self-related & SIRL - A      & N/A        & N/A        & N/A        & $97.0\%$ & $95.4\%$ & $95.3\%$ \\
		Applied                  & None    & Applied & Self-related & SIRL - WS     & N/A        & N/A        & N/A        & $85.7\%$ & $87.8\%$ & $88.7\%$ \\
		\bottomrule[1.1pt]
	\end{tabular}
\end{table*}

We further test the well-trained neural network modules using the above-mentioned five methods in the tasks to form shapes ``1" and ``2." The final similarity and the number of involved agents in each task are listed in TABLE \uppercase\expandafter{\romannumeral3}, where each value is an average of five replications. Note that the task complexity is largely related to the number of agents involved. With an increase in task complexity, the number of iterations required to reach convergence also increases, whereas the final similarity normally declines. In the first five rows of TABLE \uppercase\expandafter{\romannumeral3}, we can observe that the neural network modules fully trained in the task to form shape ``4" can also be used to form other shapes (i.e., ``1" and ``2"), and the methods with stigmergy yield superior portability to those without it.

\begin{figure}
	\centering
	\includegraphics[width=0.4\textwidth]{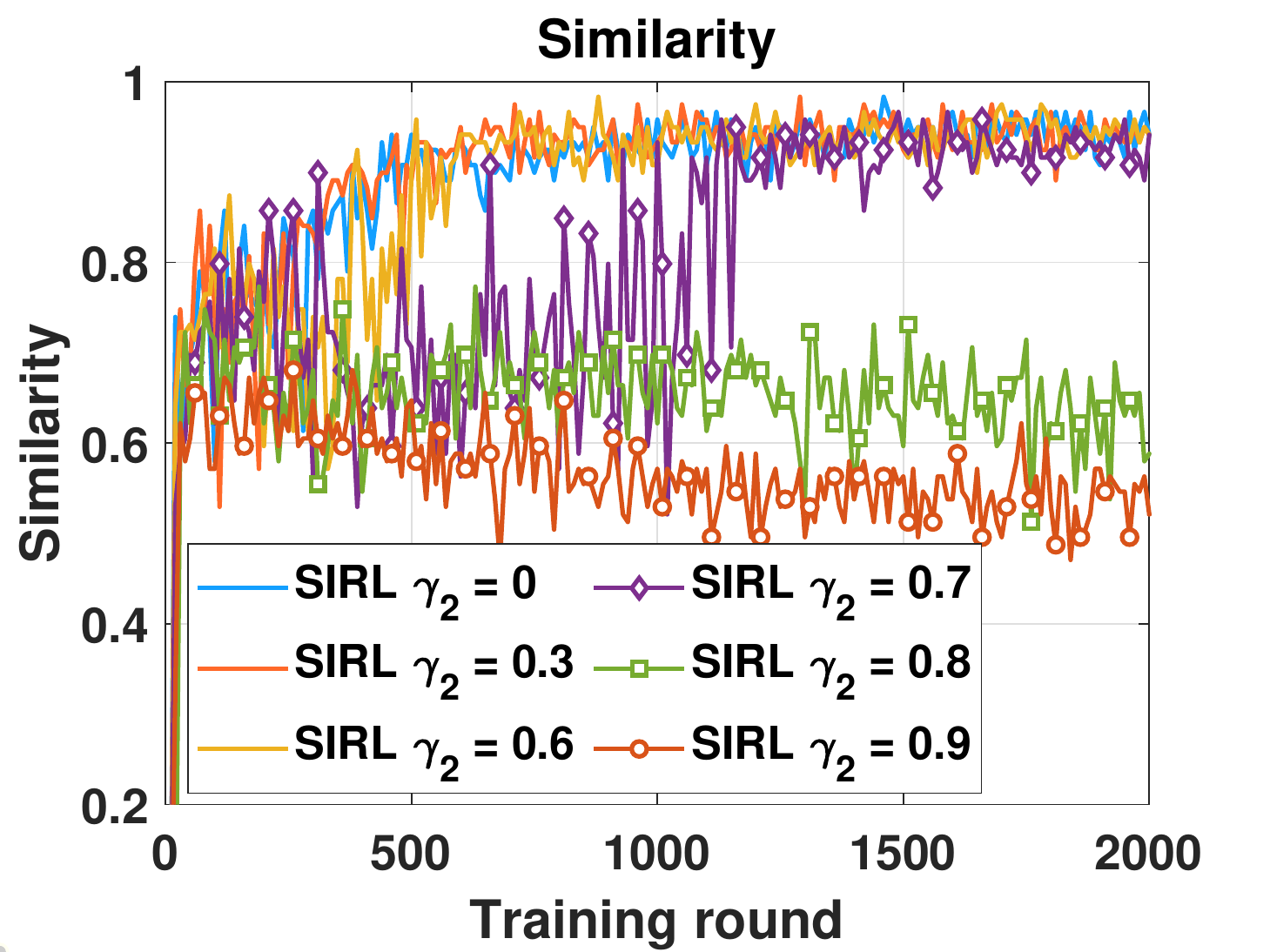}	
	\caption{Training performance of stigmergic independent reinforcement learning (SIRL) when $\gamma_2$ takes different values.}
	\label{fig7}
\end{figure}

\begin{figure}
	\centering
	\includegraphics[width=0.4\textwidth]{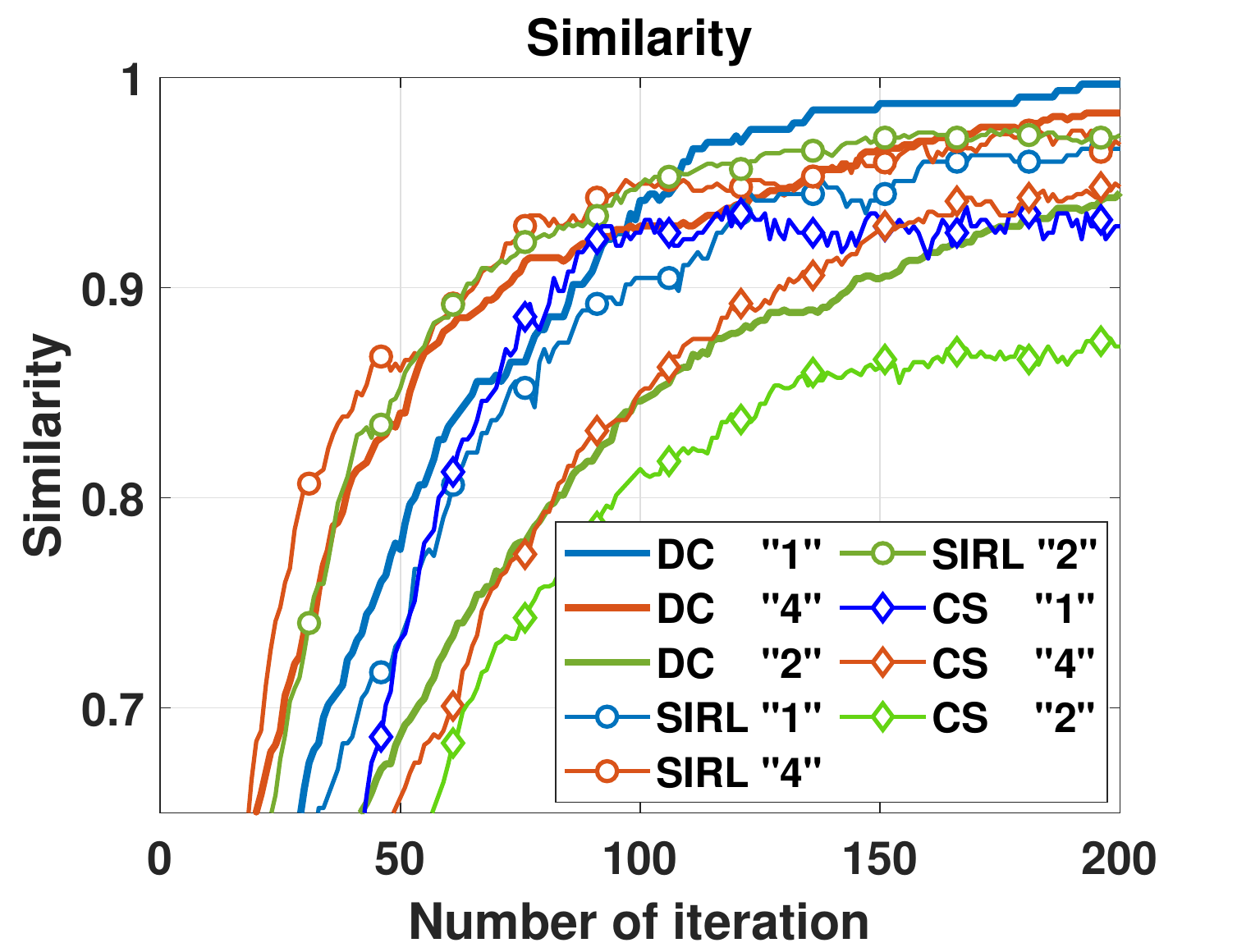}	
	\caption{Performance comparison between centralized selection (CS), decentralized (DC), and stigmergic independent reinforcement learning (SIRL) in three different tasks.}
	\label{fig8}
\end{figure}

\begin{figure*}
	\centering	
	\subfigure[]{%
		\includegraphics[width=0.33\textwidth]{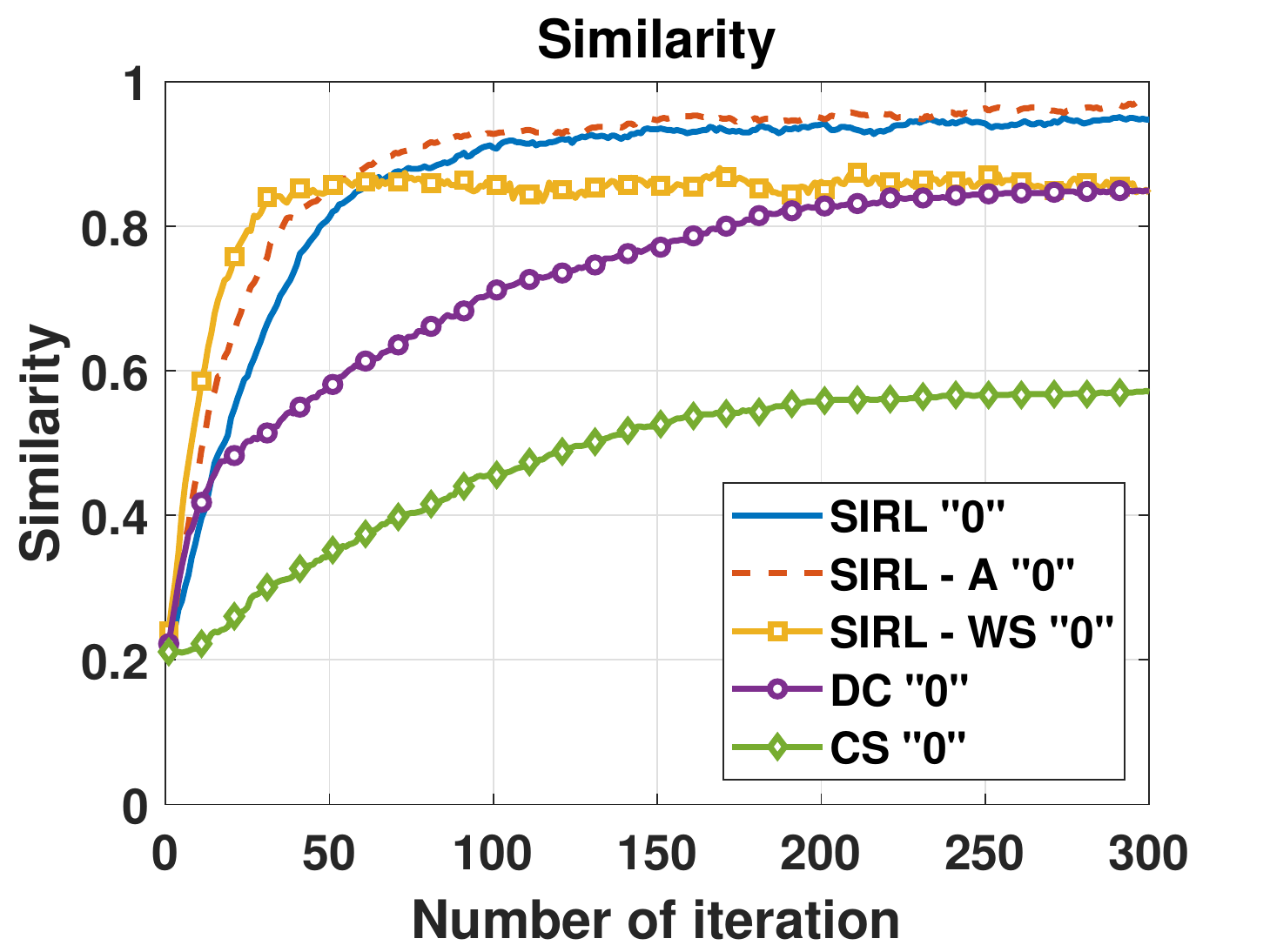}}%
	\subfigure[]{%
		\includegraphics[width=0.33\textwidth]{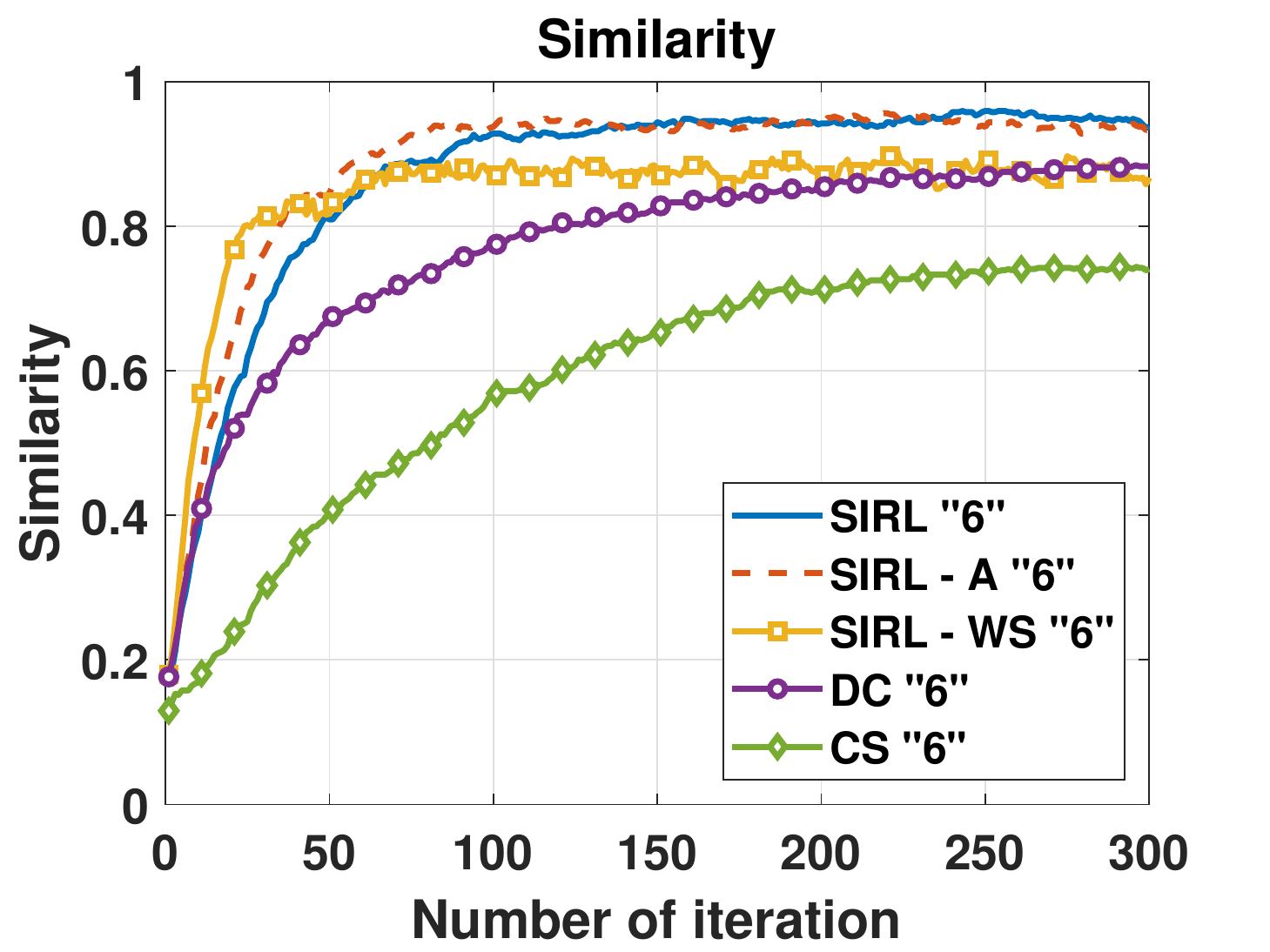}}%
	\subfigure[]{%
		\includegraphics[width=0.33\textwidth]{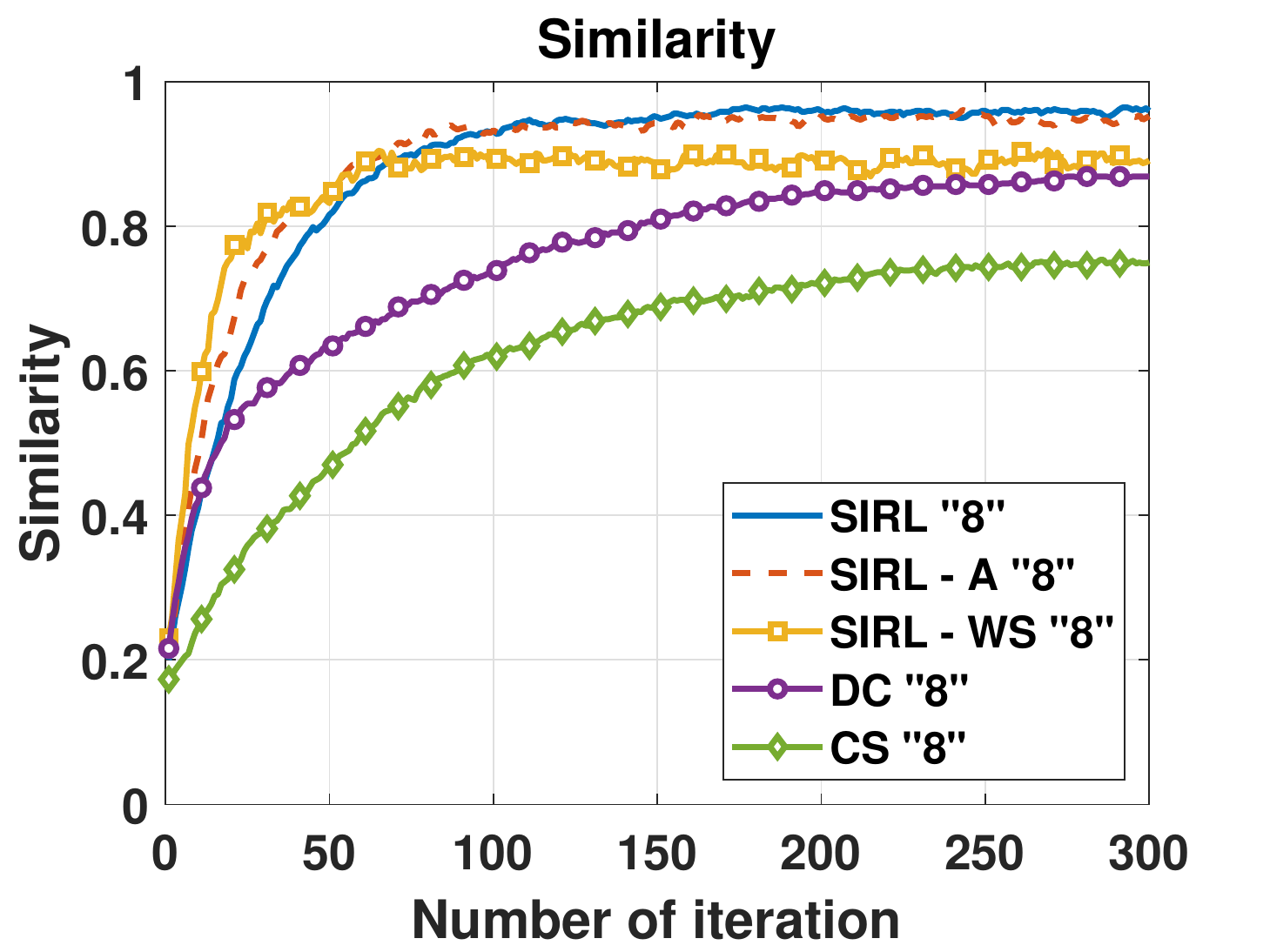}}%
	\caption{Performance comparison using different methods in the tasks to form shapes ``0," ``6," and ``8."}
	\label{fig9}
\end{figure*}

\begin{table}[htbp]
	\centering
	\caption{Manually Set Received Reward for the Decentralized (DC) Method.}
	\label{tb4}
	\begin{tabular}{c|c|c|c|c|c|c|c}
		\toprule[1.1pt]
		Number of neighbors          & 4    & 3  & 3   & 2   & 1  & 1  & 0 \\
		\hline                       
		In the labeled area          & $*$  & 1  & 0   & $*$ & 1  & 0  & $*$ \\
		\hline
		Reward                       & 0    & 4  & 12  & 8   & 8  & 12 & 12 \\
		\bottomrule[1.1pt]
	\end{tabular}
\end{table}

\begin{figure}
	\centering
	\includegraphics[width=0.38\textwidth]{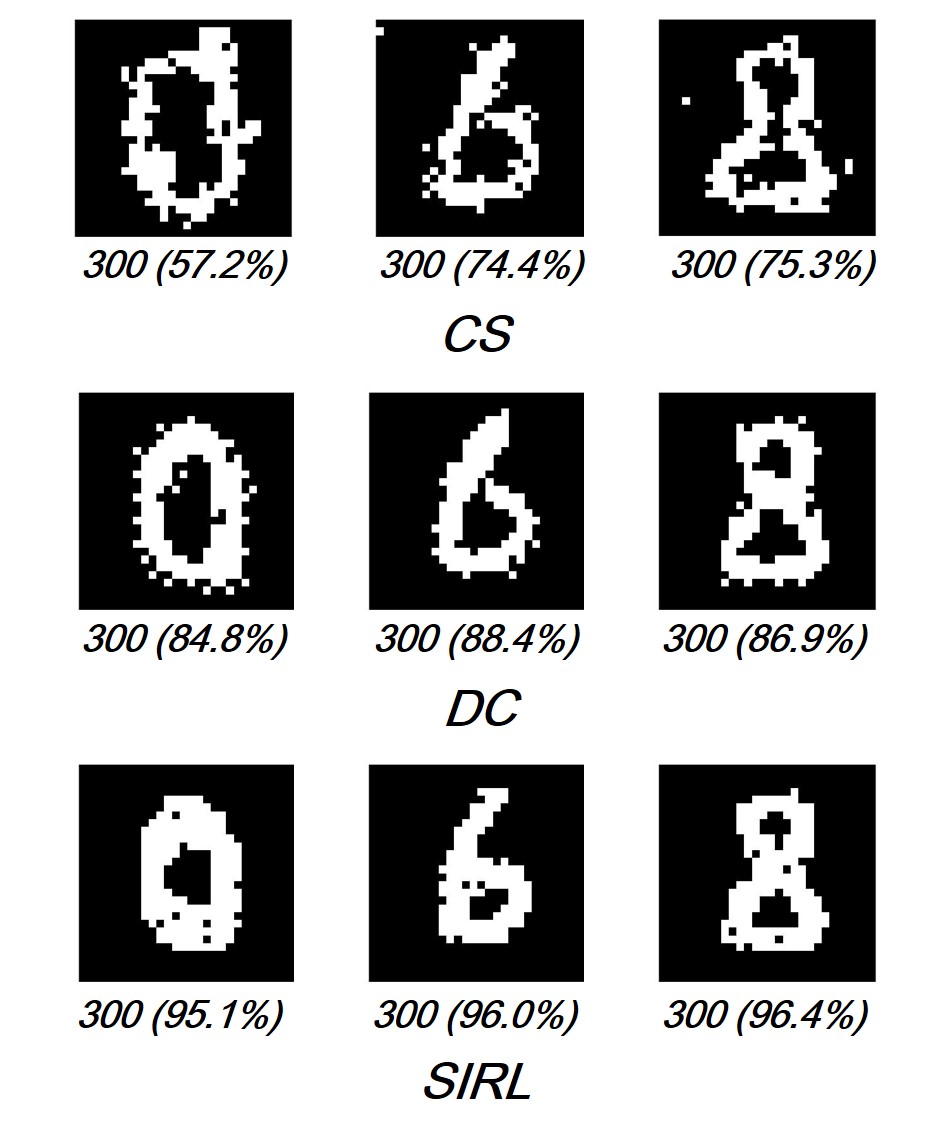}	
	\caption{Final shapes formed by centralized selection (CS), decentralized (DC), and stigmergic independent reinforcement learning (SIRL).}
	\label{fig10}
\end{figure}
 
Next, we compare the performance with the centralized selection (CS) method presented in \cite{Xing2019Brain}, in which a certain number of agents are selected synchronously at each iteration through a threshold-based function\cite{Dorigo2000Ant} to perform actions, while the other agents stop. Furthermore, we convert the CS method into a decentralized method called DC, in which each agent can automatically decide to perform an action or not through comparison with surrounding neighbors. Thus, the number of active agents may vary within each task or iteration. Instead of competing based on the action priority, as in SIRL, agents in DC can directly determine their moving priorities in terms of the received rewards. In particular, an agent can move in the next time step $t+1$ only when it receives the maximum reward in time step $t$ within the comparison range, which has the same size as the Moore neighborhood. As illustrated in TABLE \uppercase\expandafter{\romannumeral4}, the reward received by each agent in DC is carefully tuned and closely related to its surrounding conditions. Here, each agent should consider the existences of up to four nearest neighbors in the Up, Down, Left, and Right directions. The label ``1" (or ``0") in the second line signifies that this agent is in the labeled (or unlabeled) area, whereas ``*" indicates to ignore this condition. We observe that agents in the unlabeled area appear to receive greater rewards and thus obtain more action opportunities. In addition, similar to CS, each agent in DC is also designed to approach its selected attractor in a circular path so as to make most agents in the unlabeled area move around the labeled area to accelerate the convergence \cite{Xing2019Brain}.

In Fig. \ref{fig8}, we present a performance comparison between CS, DC, and SIRL in the task of forming three different shapes (i.e., ``1," ``4," and ``2").  Their final similarities are listed in the first, sixth, and seventh rows of TABLE \uppercase\expandafter{\romannumeral3}. We can observe that the performance of CS and DC tends to decline sharply with an increase in task complexity and the number of involved agents. Moreover, in CS and DC, the value of received individual rewards and the moving manner of each agent must be determined manually, which is impractical in more complex scenarios. In contrast, the performance of SIRL reaches a level comparable to that of CS or DC, and achieves a superior convergence rate in more complex tasks, such as the task to form the shape ``2."

Next, we increase the task complexity and add a ring structure into the shape to be formed so as to give it a more complex topology. The performance comparison between CS, DC, and SIRL in the tasks to form shapes ``0," ``6," and ``8" is provided in Fig. \ref{fig9}. The numerical results are also listed in the first, sixth, and seventh rows of TABLE \uppercase\expandafter{\romannumeral3}. The neural network modules utilized in SIRL are still fully trained in the previous task to form shape ``4." In Fig. \ref{fig9}, there is a large performance difference between CS or DC and SIRL. The reason is that it is generally easier for agents in the unlabeled area to win the moving opportunities; however, they move around the labeled area repeatedly in CS or DC. Consistent with our previous discussion, both the CS and DC methods perform better in less complex scenarios, such as the task to form the shape ``1." However, they reach a bottleneck for complex shapes that contain ring structures because agents in the unlabeled area are blocked by other agents located at the edge of the labeled area. 

We present the final shapes formed by the three methods in Fig. \ref{fig10}. It can be clearly observed that many agents in CS or DC are detained on the periphery of the labeled area, ultimately reducing the final similarity. In contrast, the performance of SIRL remains stable regardless of the task complexity, which benefits from the learning process of the federal training method. Compared with DC, the number of agents that are selected at each iteration must be predetermined in CS. However, the most appropriate number may depend on the specific task or iteration. Thus, CS would obtain lower final similarities or require a larger number of total steps in different tasks. In addition, the performance of CS also declines dramatically when the shape to be formed contains a ring structure. This phenomenon can also be observed in DC, as the two methods share the same manner of motion.

We further test the effect of different ranges of the coordination channel of the conflict-avoidance mechanism on the final performance based on SIRL. Instead of relying on the Moore neighbors, in the SIRL-A method, we reduce the maximum range of the coordination channel to $1$; thus, only four neighbors are added in the comparison of action priority. Furthermore, in the SIRL-WS method, we disable the conflict-avoidance mechanism (i.e., well-trained Evaluation Module) and provide each agent with the action opportunity at each time step, which can be regarded as the maximum range of the coordination channel being set to $0$. The performance of the two methods is illustrated in Fig. \ref{fig9}. The numerical results are also listed in the last two rows of TABLE \uppercase\expandafter{\romannumeral3}. It can be seen that the performance of SIRL-A reaches a similar level to that of SIRL, and achieves an even higher convergence rate, since more agents can obtain action opportunities and participate in the task at an early stage. In addition, the curves of SIRL-WS in various tasks grow faster but stop at similarly lower levels. It can be concluded that the conflict-avoidance mechanism plays an important role in multi-agent collaboration and can reduce the behavioral localities of agents and achieve a higher final similarity. 

\begin{figure}
	\centering
	\includegraphics[width=0.4\textwidth]{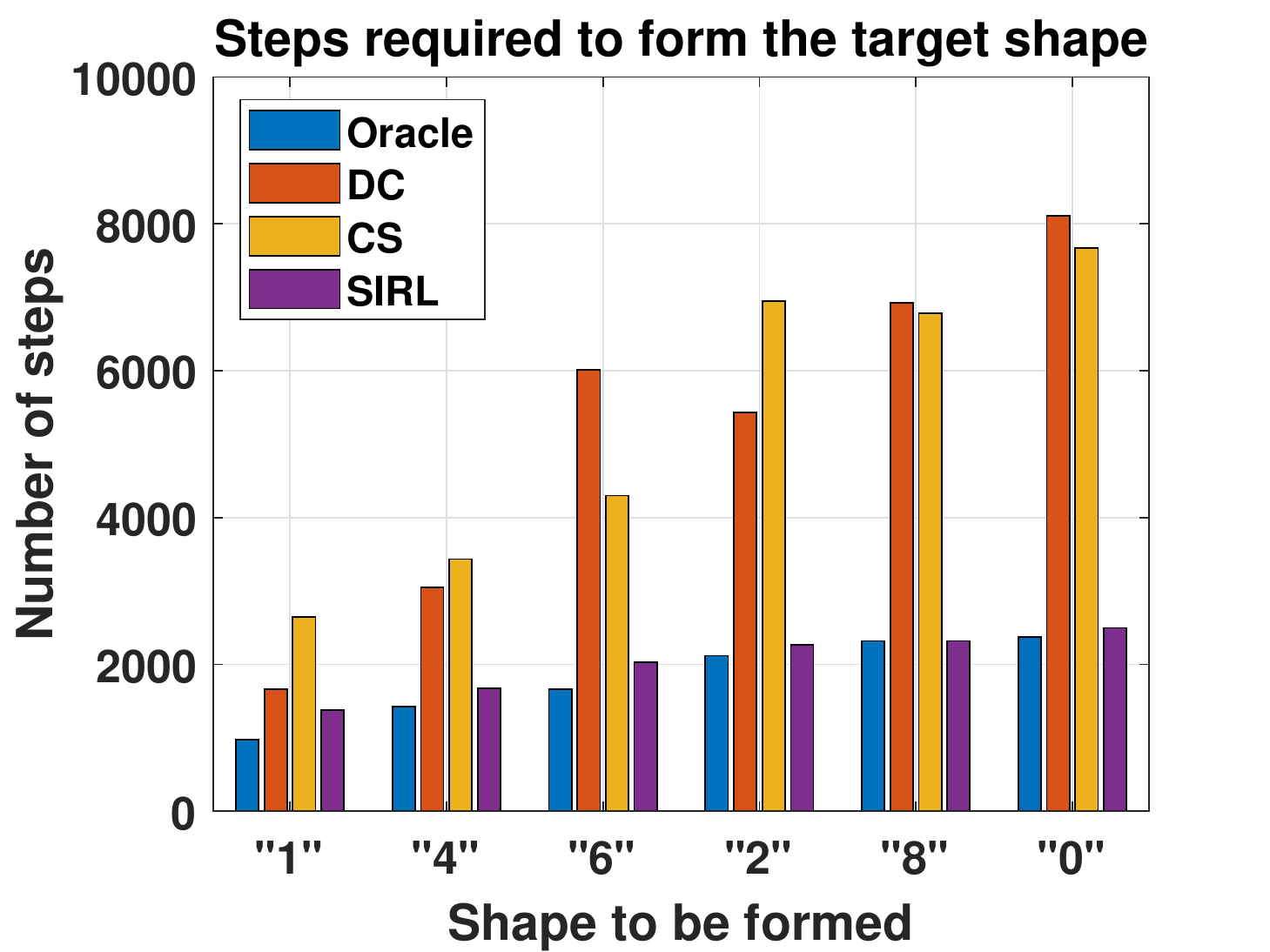}	
	\caption{Number of steps required using different methods.}
	\label{fig11}
\end{figure}

As an important metric, the number of total steps required by all involved agents to form the target shape should be especially considered. In Fig. \ref{fig11}, we present the total number of steps when the performance reaches convergence using four different methods. In the Oracle method, it is assumed that the location of each agent and the target shape are known in advance. Therefore, for each vacancy in the labeled area, the nearest agent is moved in sequence to fill it greedily. This scheme represents a simple but effective control method in the extraordinary case in which all information is known, even though the greedy algorithm does not necessarily produce the optimal result. In Fig. \ref{fig11}, the formed shapes are arranged from small to large based on the number of involved agents. There is a trend that as the number of involved agents increases, the number of total steps required also increases. We can observe that the performance of SIRL reaches a level comparable to that of Oracle. Moreover, since the agents are controlled in parallel, SIRL may spend less time completing tasks in real scenarios. 

\section{Conclusions}
This study utilizes the advantages of the stigmergy mechanism rooted in a natural MAS and contributes to integrating stigmergic collaboration into the DRL-based decision-making process of each involved intelligent agent. In particular, in our proposed SIRL algorithm, various agents are coordinated through the stigmergic functionality and conflict-avoidance mechanism in fulfilling the corresponding tasks. As an enhancement to stigmergy, the proposed conflict-avoidance mechanism can further improve the coordination performance, especially when agents do not fully cooperate. In addition, the RL process of each agent is strengthened through the proposed federal training method. Compared with a traditional distributed learning algorithm, our learning method allows agents to optimize their internal neural networks while maintaining their external interaction processes required for effective inter-agent collaboration. Furthermore, due to the introduction of the stigmergic RL process, our proposed SIRL scheme can be further applied to more complex cooperative tasks for which the agent's behavioral policy cannot be predetermined. The results of numerical simulations verify the effectiveness of our proposed scheme with significant improvements in both efficiency and scalability.

In future work, we plan to implement the proposed SIRL scheme in real-world application scenarios. As a preliminary step, our scheme has recently been implemented on multiple mobile robots in team formation tasks, which are similar to those presented in this paper \cite{KunDemo2020}, and achieve satisfactory results. We also plan to investigate the performance of the SIRL scheme in other coordination scenarios, such as the control of autonomous vehicles in intelligent urban transportation. In addition, determining how to improve the network training efficiency while considering the communication delay and cost among various agents is also a valuable direction that we will consider in future research.

\vspace{0.5cm}
\balance
\bibliographystyle{IEEEtran}
\bibliography{stigmergy}

\end{document}